\documentclass[letterpaper]{article} 
\usepackage{aaai23-r2hcai}  
\usepackage{times}  
\usepackage{helvet}  
\usepackage{courier}  
\usepackage[hyphens]{url}  
\usepackage{graphicx} 
\usepackage{natbib}  
\usepackage{caption} 
\usepackage{amsfonts}
\usepackage{amssymb}
\usepackage{amsmath,upgreek}
\usepackage{titlesec}
\usepackage{cite}
\usepackage{amsfonts,amssymb}
\usepackage{multirow}
\usepackage{multicol}
\usepackage{ulem}
\usepackage{graphicx}
\usepackage{subfigure}
\usepackage{subfig}
\usepackage{subcaption}
\usepackage{algorithm}
\usepackage{algorithmic}
\usepackage{booktabs}
\usepackage{multirow}
\usepackage{threeparttable}
\usepackage{bm}
\usepackage{xcolor}

\usepackage{newfloat}
\usepackage{listings}
\DeclareCaptionStyle{ruled}{labelfont=normalfont,labelsep=colon,strut=off} 
\floatstyle{ruled}
\newfloat{listing}{tb}{lst}{}
\floatname{listing}{Listing}
\usepackage{fancyhdr}
\title{Bi-Mamba+: Bidirectional Mamba for Time Series Forecasting}
\author{
    Aobo Liang\textsuperscript{1} \quad
    Xingguo Jiang\textsuperscript{1} \quad
    Yan Sun\textsuperscript{1*} \quad
    Xiaohou Shi\textsuperscript{2} \quad
    Ke Li\textsuperscript{3}
}
\affiliations {
    \textsuperscript{1} 
    Beijing University of Posts and Telecommunications, China \\
    \textsuperscript{2} 
    China Telecom Research Institute, Beijing, China \\
    \textsuperscript{3} 
    China Telecom Corporation Intelligent Cloud Network Operation Center, Beijing, China \\
    \{liangaobo, sunyan\}@bupt.edu.cn, jxgaugusto@gmail.com,\\ \{shixh6, lik24\}@chinatelecom.cn
}

\usepackage{bibentry}

\begin{document}
\maketitle

\begin{abstract}
Long-term time series forecasting (LTSF) provides longer insights into future trends and patterns. Over the past few years, deep learning models especially Transformers have achieved advanced performance in LTSF tasks. 
However, LTSF faces inherent challenges such as long-term dependencies capturing and sparse semantic characteristics.
Recently, a new state space model (SSM) named Mamba is proposed. With the selective capability on input data and the hardware-aware parallel computing algorithm, Mamba has shown great potential in balancing predicting performance and computational efficiency compared to Transformers. To enhance Mamba's ability to preserve historical information in a longer range, we design a novel Mamba+ block by adding a forget gate inside Mamba to selectively combine the new features with the historical features in a complementary manner.
Furthermore, we apply Mamba+ both forward and backward and propose Bi-Mamba+, aiming to promote the model's ability to capture interactions among time series elements.
Additionally, multivariate time series data in different scenarios may exhibit varying emphasis on intra- or inter-series dependencies. Therefore, we propose a series-relation-aware decider that controls the utilization of channel-independent or channel-mixing tokenization strategy for specific datasets. 
Extensive experiments on 8 real-world datasets show that our model achieves better predictions compared with state-of-the-art methods.
Our code is available at https://github.com/Leopold2333/Bi-Mamba+.

\end{abstract}

\section{Introduction}
Time series forecasting (TSF) is an indispensable part of many fields such as traffic flow prediction\citep{guo2019attention}, energy management\citep{uremovic2022new}, weather forecasting\citep{zhang2022solar}, finance\citep{sezer2020financial}, etc. Among TSF tasks, the long-term time series forecasting (LTSF) task predicts the trend, periodicity and other key patterns of data with a longer future observations, enabling better long-term strategies development, resource allocation planning and risk management.

In the past few years, a large number of deep learning models have been developed for LTSF tasks. Among these efforts, Transformer-based models achieve great success. 
Transformers use self-attention mechanism to capture long-term dependencies of the time series. In addition, the vanilla Transformer implicitly models the inter-series dependencies through channel-mixing embeddings. However, the quadratic complexity of the self-attention mechanism consumes excessive computational resources, resulting in slow training and inference speeds. Although many works like Informer\citep{informer} and Autoformer\citep{autoformer} try to propose some sparse attention to solve this issue in recent years, they face the challenges of balancing computational efficiency and predicting performance. Moreover, these models do not explicitly capture the inter-series dependencies, which may cause inadequate modeling of the relationships between different time series\citep{zhang2024sageformer}.

Recently, state-space models (SSM) emerge as a promising architecture for sequence modeling\citep{gu2021combining,gu2021efficiently}. Among them, a recent study, Mamba\citep{gu2023mamba}, has achieved remarkable results in sequence processing tasks such as natural language processing\citep{lieber2024jamba}, DNA sequence learning, audio waveform processing\citep{gu2023mamba} and computer vision\citep{zhu2024vision}. Benefiting from its design of selective scanning, Mamba shows superior performance for long sequence modeling, which makes it potentially suitable for the LTSF task. However, there are limited utilizations of SSM in LTSF currently, which may stem from the inherent challenges in time series analysis tasks as follows:

\begin{itemize}
    \item \textbf{Long-term time series modeling}. 
    LTSF deals with larger magnitudes of sequence elements and is often more affected by data non-stationarity, noise and outliers, making it more difficult to capture long-term dependencies\citep{liu2022non}. Furthermore, the semantic information density of time series data at time points is lower than other types of sequence data. Directly modeling point-wise tokens may introduce redundant noise\citep{cao2023inparformer}. Works such as PatchTST\citep{patchtst} and Crossformer\citep{crossformer} emphasis capturing richer semantic information by dividing time series into patches for tokenization. Compared with point-wise input tokens, patch-wise tokens reduces the number of sequence elements and leads to lower computational complexity. A recent work iTransformer\citep{liu2023itransformer} uses a simple fully connected layer to map the whole sequence to hidden states. Although iTransformer gets state-of-the-art (SOTA) performance, it is coarse-grained and is not conducive to capturing fine-grained evolutionary patterns inside the time series. Therefore, we are motivated to model the time series in a patching manner.
    \item \textbf{Emphasis on intra- or inter-series dependencies}. 
    Time series data may have multiple variables with complex correlations between each other. 
    For example, the traffic flow of interconnected roads within a transportation network influence each other in different manners during different periods of time. 
    However, capturing inter-series dependencies using channel-mixing strategies does not always produce SOTA performance. Works such as PatchTST and NHITS\citep{challu2023nhits} get superior performance by capturing intra-series dependencies only. These models process the multivariate time series (MTS) data in a channel-independent way, where each univariate series is input to the model independently to enhance the robustness of the model. It can be inferred that different datasets shows different emphasis on intra- or inter-sequence dependencies.
    A recent study TimeMachine\citep{ahamed2024timemachine} proposes a unified structure for channel-independent and channel-mixing tokenization strategies. It aims to handle both intra-series-emphasis and inter-series-emphasis situations. However, the boundary for the selection of tokenization strategies is ambiguous and the statistical characteristics of datasets are overlooked.
\end{itemize}

To solve the above challenges, 
we design an improved Mamba block named Mamba+ through adding a forget gate in Mamba to selectively combine the new features with the historical features in a complementary manner, therefore preserving historical information in a longer range. 
Based on the Mamba+, we propose \textbf{Bi}directional \textbf{Mamba+} (Bi-Mamba+) to model the MTS data from both forward and backward, enhancing the model's robustness and ability to capture interactions between time series elements.
In addition, to address the varying emphasis on intra-series evolutionary patterns and inter-series interactions, we design a \textbf{S}eries-\textbf{R}elation-\textbf{A}ware (SRA) decider inspired from Spearman coefficient correlation. 
The SRA decider measures the proportion of highly correlated series pairs in the MTS data to automatically choose channel-independent or channel-mixing tokenization strategies. 
Furthermore, we divide the input sequence into patches and generate patch-wise tokens based on the chosen tokenization strategy. The patch-wise tokens contain richer semantic information and encourage the model to learn the long-term dependencies of the time series in a finer granularity.
The main contributions of this paper are summarized as follows:

\begin{itemize}
    \item We propose Bi-Mamba+ for LTSF task. We design an improved Mamba+ block and model the MTS data from both forward and backward to provide more accurate predictions.
    \item We design an SRA decider based on the Spearman correlation coefficient to automatically choose channel-independent or channel-mixing tokenization strategies. Furthermore, we divides the time series into patches to capture long-term dependencies in a finer granularity.
    \item We conduct extensive experiments on 8 real-world datasets that are widely used. The results show that Bi-Mamba+ achieves superior performance with varying prediction lengths.
\end{itemize}

\section{Related Work}
\subsection{Time Series Forecasting}
TSF aims to predict future values based on historical observations\citep{lim2021time}. Recently, models based on deep neural network achieve great success in TSF. Among these deep learning models, Transformer-based models\citep{vaswani2017attention} get outstanding performance because of the self-attention mechanism. However, the quadratic complexity to the length of the sequence limits these models’ application on long-term time
series forecasting. Therefore, researches in recent years focus on balancing the computing efficiency and predicting performance of Transformers. Informer\citep{informer} proposes a \textit{ProbSparse} mechanism which selects top-k elements of the attention weight matrix to make distillation operation on self-attention. Autoformer\citep{autoformer} uses time series decomposition and proposes an \textit{Auto-Correlation} mechanism inspired by the stochastic process theory. Pyraformer\citep{liu2021pyraformer} introduces the pyramidal attention module to summarizes features at different resolutions and model the temporal dependencies of different ranges. FEDformer\citep{zhou2022fedformer} develops a frequency enhanced Transformer through frequency domain mapping. PatchTST\citep{patchtst} divides each univariate sequence into patches and uses patch-wise self-attention to model temporal dependencies. Crossformer\citep{crossformer} adopts a similar patching operation but additionally employs a Cross-Dimension
attention to capture inter-series dependencies. While patching helps reduce the number of sequence elements to be processed and extract richer semantic information, the self-attention layers are only used on the simplified sequences. Therefore, the performance bottlenecks still occur when these models deal with longer sequences. To tackle this issue, iTransformer\citep{liu2023itransformer} inverts the attention layers to straightly model inter-series dependencies. However, the tokenization approach adopted by iTransformer is simply passing the whole sequence through a Multilayer Perceptron (MLP) layer, which overlooks the complex evolutionary patterns inside the time series. Overall, Transformer-based models still face the challenges in computational efficiency and predicting performance.

\begin{figure*}[t]
    \centering
    \includegraphics[width=1.0\textwidth]{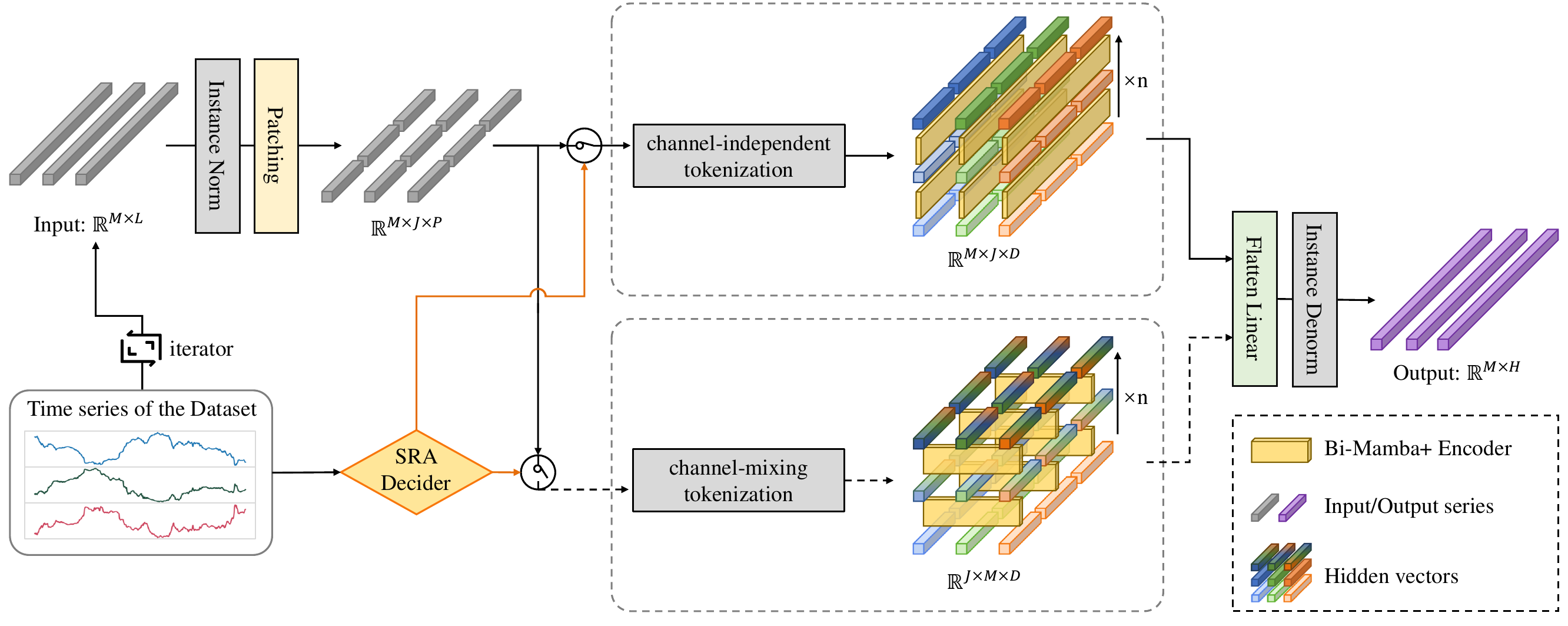}
    \caption{The architecture of Bi-Mamba+. The input time series is first divided into patches and then pass through a channel-independent or channel-mixing embedding layer based on the data characteristics of specific datasets. The embeddings are then fed into multiple Bi-Mamba Encoders and get the final output through an MLP projector.}
    \label{fig:archetecture}
\end{figure*}

\subsection{SSM-based models}
Deep learning models for time series forecasting are initially based on Recurrent Neural Network (RNN) or Convolutional Neural Network (CNN). RNNs process the sequence elements step by step and maintain a hidden state which is updated with each input element. These models are relatively simple and have excellent inference speed. However, the calculated gradient must be passed through all cells one by one, which limits the training speed and leads to  forgetting long-term information. CNNs use convolutional kernel to emphasis local information. These models benefit from parallel computing and have faster training speed. However, the convolutional calculating process limits the inference speed and overlook the long-term global information. To solve the issues of these two models, a new model structure is proposed, namely State Space Models (SSM). SSM\citep{gu2021combining,gu2021efficiently} is inspired by the continious system. It is trained in parallel like CNN and inferences fastly like RNN. Some previous works have attempted to use SSM in TSF. SSDNet\citep{lin2021ssdnet} combines
the Transformer architecture with SSM to provide probabilistic and interpretable forecasts. SPACETIME\citep{gu2021efficiently} proposes a new SSM parameterization based on the companion matrix to enhance the expressivity of the model and introduces a “closed-loop” variation of the companion SSM for long-term time series forecasting.

Recently, a new SSM-based model, Mamba\citep{gu2023mamba}, is proposed. It introduces parameterized matrices and a hardware-aware parallel computing algorithm to SSM and achieves superior performance on language modeling, DNA sequence and audio waveform processing tasks. Several Mamba-based derivative models have been developed and used for computer vision tasks\citep{zhu2024vision,ma2024u} and time series tasks. S-Mamba\citep{wang2024mamba} explores to use Mamba to capture inter-series dependencies of MTS data. It embeds each univariate time series like iTransformer and feeds the embeddings into Mamba blocks to model the relationships of different time series. However, the tokenization approach may overlook the complex evolutionary patterns inside the time series. MambaMixer\citep{behrouz2024mambamixer} adjusts the Mamba block to bidirectional and uses two improved blocks to capture inter- and intra-series dependencies simultaneously. However, the gating branch is used to filter new features of both forward and backward directions, which may cause challenges for extracting new features.
TimeMachine\citep{ahamed2024timemachine} proposes a multi-scale quadruple-Mamba architecture to unify the handling of channel-mixing and channel-independence situations. However, the channel-mixing and channel-independent strategies are chosen simply based on the length of historical observations and variable number of different datasets. Therefore, the characteristics of the MTS data are not fully considered.

\section{Methodology}
\subsection{Preliminaries}
\subsubsection{Long-term multivariate time series forecasting.}
We consider the Long-term multivariate time series forecasting task as follows: given a multivariate time series $\mathbf{X}_{in}=\left[x_1,x_2,...,x_L\right]\in\mathbb{R}^{L\times M}$, we predict the future values $\mathbf{X}_{out}=\left[x_{L+1},x_{L+2},...,x_{L+H}\right]\in\mathbb{R}^{H\times M}$, where $L$ is the length of historical look-back window, $M$ represents
the feature dimension, and $H$ is the prediction horizons.

\subsubsection{State Space Models.}\label{sec::ssm} SSM is inspired by the continious system, using first-order differential equations to map input function $x(t)$ to output function $y(t)$ through hidden state $h(t)$, defined as follows:

\begin{equation}
h^\prime(t)=\mathbf{A}h(t)+\mathbf{B}x(t),\quad y(t)=\mathbf{C}h(t)
\end{equation}
where $\mathbf{A}\in\mathbb{R}^{N\times N}$, $\mathbf{B}\in\mathbb{R}^{D\times N}$ and $\mathbf{C}\in\mathbb{R}^{N\times D}$. The variable $N$ and $D$ refer to the state expansion factor and dimension factor respectively. 
The continuous parameters $\mathbf{A,B}$ can be discretized to $\overline{\mathbf{A}}, \overline{\mathbf{B}}$ by zero-order holding and time sampling at intervals of $\Delta$, defined as follows:

\begin{equation}
\begin{aligned}
    \overline{\mathbf{A}}&=\exp{(\Delta\mathbf{A})},\\
    \overline{\mathbf{B}}&=(\Delta\mathbf{A})^{-1}(\exp{(\Delta\mathbf{A})}-\mathbf{I})\cdot\Delta\mathbf{B}.
\end{aligned}
\end{equation}
The formula of discretized SSM can then be written as:
\begin{equation}
    h_k=\overline{\mathbf{A}}h_{k-1}+\overline{\mathbf{B}}x_k,\quad y_k=\mathbf{C}h_k
\end{equation}

The discretized SSM\citep{gu2021combining} can be trained in parallel in a convolutional operation way and make efficiently inference in a recurrent neural network manner. By introducing HIPPO Matrix\citep{gu2020hippo} to the initialization of matrix $\mathbf{A}$, a variant of SSM, namely the structured state space model (S4)\citep{gu2021efficiently}, makes improvement on the ability to model long-term dependencies.

Mamba\citep{gu2023mamba} parameterizes the matrices $\mathbf{B,C}$ and $\Delta$ in a data-driven manner, introducing a selection mechanism into S4 model. In addition, Mamba uses a novel hardware-aware parallel computing algorithm to ensure the efficient training of the model. With linear computational complexity and outstanding capabilities in modeling long-term dependencies, Mamba shows great potential in time series forecasting tasks and is expected to be a competitor to Transformer-based models.

\subsection{Overview}
The overall architecture of Bi-Mamba+ is shown in Fig. \ref{fig:archetecture}. 
We first calculate the tokenization strategy indicator through the SRA decider. we then divide the input series into patches and generate patch-wise tokens based on the tokenization strategy indicator. The obtained tokens are fed into multiple Bi-Mamba+ encoders. We adopt a flatten head and a linear projector to get the final output.

\subsection{Instance Normalization}
Time series in the real world usually conform to the characteristics of non-stationary sequences\citep{adarnn,long2022multivariate}. The statistical properties of time series data usually change over time, resulting in a changing data distribution. We use RevIN \citep{revin} to eliminate the non-stationary statistics in the input sequence. The RevIn module normalizes the input batch data and denormalizs the output of the model. The instance normalization process corresponds to the \textbf{Instance Norm} and \textbf{Instance Denorm} modules in Fig. \ref{fig:archetecture}.

\begin{algorithm}[t]
\caption{SRA Decider}\label{alg:alg1}
\renewcommand{\algorithmicrequire}{\textbf{Input:}}
\renewcommand{\algorithmicensure}{\textbf{Output:}}
\begin{algorithmic}[1]
\REQUIRE Training set $T=\{t^1,t^2,...,t^M\}$ of MTS data, each series has $n$ observations
\ENSURE Tokenization strategy $ts\in \{0,1\}$
\FOR{$i=1$ to $M$}
\STATE $\rho_{i, i}=0$
\FOR{$j=i+1$ to $M$}
\STATE $\rho_{i, j}=\rho_{j, i}=1-\frac{6\sum_{k=0}^{n}\left(\mathrm{Rank}(t_k^i)-\mathrm{Rank}(t_k^j)\right)^2}{n(n^2-1)}$
\ENDFOR
\ENDFOR
\STATE initialize list $K^\lambda$ and $K^0$ of length $M$ with all elements set to 0
\FOR{$i=1$ to $M$}
\FOR{$j=1$ to $M$}
\IF{$\rho_{i,j}\ge\lambda$}
\STATE $K_i^\lambda=K_i^\lambda+1$
\ELSIF{$\rho_{i,j}\ge0$}
\STATE $K_i^0=K_i^0+1$
\ENDIF
\ENDFOR
\ENDFOR
\STATE $\rho_{\max}^\lambda=\mathrm{argMax}(K^\lambda)$, $\rho_{\max}^0=\mathrm{argMax}(K^0)$
\STATE $r=\rho_{\max}^\lambda/\rho_{\max}^0$
\IF{$r\ge 1-\lambda$}
\STATE \textbf{return}  $ts=1$ for channel-mixing tokenization
\ELSE
\STATE \textbf{return}  $ts=0$ for channel-independent tokenization
\ENDIF
\end{algorithmic}
\end{algorithm}

\subsection{Token Generalization}
\subsubsection{SRA Decider.}
Recent studies\citep{patchtst,zhou2022fedformer} show that both channel-independent and channel-mixing strategies can achieve SOTA accuracy in specific tasks. Typically, models using the channel-independent strategy can achieve better performance on datasets that have relatively few variables, while models using the channel-mixing strategy are more suitable for datasets with more variables. This can be viewed as a balance between the emphasis on intra-series dependencies and inter-series dependencies of MTS data. 

Therefore, we design a SRA decider to automatically control the tokenization process of the model. The workflow of the dicision maker is shown in Alg.\ref{alg:alg1}. For a specific dataset, we extract the training set data $T=\{t^1,t^2,...,t^M\}$ and calculate the Spearman correlation coefficients of different series $t^i$ and $t^j$, denoted as $\rho_{i,j}$, where $i$ and $j$ are the indexes of the series ranging from $1$ to $M$ and $i\ne j$. We then use threshold $\lambda$ and $0$ to filter out series pairs with positive correlation. Finally, we count the maximum number of relevant series $\rho_{\max}^\lambda$ and $\rho_{\max}^0$ in the training set and calculate the relation ratio $r=\rho_{\max}^\lambda/\rho_{\max}^0$. We adopt channel-mixing strategy to generate sequence tokens for datasets with $r\ge 1-\lambda$, otherwise, we adopt channel-independent strategy. Spearman coefficient is a nonparametric statistical indicator for evaluating the monotonic relationship between two sequences. It calculates the correlation based on the rank of the sequence elements, making it more suitable for non-linear relationships and more robust to outliers. Given two sequences $t^i$ and $t^j$, it first sorts the sequence and then define the Spearman coefficient as $\rho_{i,j}=1-\frac{6\sum_{k=0}^{n}\left(\mathrm{Rank}(t_k^i)-\mathrm{Rank}(t_k^j)\right)^2}{n(n^2-1)}$, where $n$ is the number of observation samples of each sequence and $\mathrm{Rank}(t_k^i)$ represents the rank level of the $k$-th element in the specific time series $t^i$.

\subsubsection{Tokenization Process.}
We generalize patch-wise tokens to emphasize capturing local evolutionary patterns of the time series.
Specifically, each channel univariate sequence $x^{i}_{1:L}$
is first divided into
patches $p^i\in \mathbb{R}^{J\times P}$ where $J$ is the total number of patches and $P$ is the length of each patch. We denote $S$ as the stride among patches and $J=\lceil\frac{L-P}{S}+1\rceil$. 
For channel-independent strategy, each univariate channel is concatenated to the tokens $\mathbb{E}_{ind}\in\mathbb{R}^{M\times J\times D}$, where $D$ is the hidden state dimension. For channel-mixing strategy, we first group patches with the same index of different series and pass each group through the tokenization layer. The obtained tokens can be represented as $\mathbb{E}_{mix}\in\mathbb{R}^{J\times M\times D}$.

\begin{algorithm}[t]
\caption{Implementation of Mamba+ block}\label{alg:alg2}
\renewcommand{\algorithmicrequire}{\textbf{Input:}}
\renewcommand{\algorithmicensure}{\textbf{Output:}}
\begin{algorithmic}[1]
\REQUIRE Patch-wise token sequence $\mathbb{E}_x:(B,W,D)$
\ENSURE Output features $\mathbb{E}_y:(B,W,E)$
\STATE $\mathrm{x}:(B,W,E)\leftarrow\mathrm{Linear^x}(\mathbb{E}_x)$
\STATE $\mathrm{z}:(B,W,E)\leftarrow\mathrm{Linear^z}(\mathbb{E}_x)$
\STATE $\mathrm{x}^\prime:(B,W,E)\leftarrow\mathrm{SiLU(Conv1D(x))}$
\STATE $\mathbf{A}:(E,N)\leftarrow\mathrm{Parameter}^\mathbf{A}$
\STATE $\mathbf{B}:(B,W,N)\leftarrow\mathrm{Linear}^\mathbf{B}(\mathrm{x}^\prime)$
\STATE $\mathbf{C}:(B,W,N)\leftarrow\mathrm{Linear}^\mathbf{C}(\mathrm{x}^\prime)$
\STATE $\Delta:(B,W,E)\leftarrow\log(1+\exp(\mathrm{Linear}^\Delta(\mathrm{x}^\prime))+\mathrm{Parameter}^\Delta)$
\STATE $\overline{\mathbf{A}},\overline{\mathbf{B}}:(B,W,E,N)\leftarrow\mathrm{discretize(\Delta,\mathbf{A},\mathbf{B})}$
\STATE $\mathrm{y}:(B,W,E)\leftarrow\mathrm{SSM}(\overline{\mathbf{A}},\overline{\mathbf{B}},\mathbf{C})(\mathrm{x}^\prime)$
\STATE $\mathrm{y}^\prime:(B,W,E)\leftarrow\mathrm{y}\otimes\mathrm{SiLU(z)}+\mathrm{x}^\prime\otimes(1-\sigma(\mathrm{z}))$
\STATE $\mathbb{E}_y:(B,W,D)\leftarrow\mathrm{Linear^{y^\prime}(y^\prime)}$
\STATE \textbf{return} $\mathbb{E}_y$
\end{algorithmic}
\label{alg1}
\end{algorithm}

\subsection{Mamba+ Block}
An original Mamba block uses two branches to process the input features. One of the branches, denoted as $b_1$, passes the input features through a 1-D convolutional module and a SSM block. The other branch, denoted as $b_2$, feeds the input features into a SiLU activation function to serve as a gate. Although the HIPPO matrix\citep{gu2021efficiently} embedded in the SSM block can retain a fairly long-term historical information, the obtained result is filtered directly through the gate of another branch, resulting in the tendency to prioritize proximal information\citep{wang2024mamba}. Therefore, we propose an improved Mamba+ block specifically designed for LTSF. Particularly, we add a forget gate calculated by $gate_{f}=1-gate_{b_2}$, where $gate_f$ represents the forget gate and $gate_{b_2}$ is the result of sigmoid function in $b_2$. The output of the 1-D convolutional module $x^\prime$ is multiplied with $gate_{f}$ and then added to the filtered result of SSM. $gate_{f}$ and $gate_{b_2}$ selectively combine the added new features with the forgotten historical features in a complementary manner. 
Compared to directly filtering the obtained new features, Mamba+ emphasis preserving historical information in a longer range. The detailed flow of Mamba+ is shown in Alg.\ref{alg:alg2}. The reprogrammed forward and backward propagation process of the hardware-aware parallel computing algorithm is shown in Appendix.

\begin{algorithm}[t]
\caption{Implementation of Bi-Mamba+ Encoder}\label{alg:alg3}
\renewcommand{\algorithmicrequire}{\textbf{Input:}}
\renewcommand{\algorithmicensure}{\textbf{Output:}}
\begin{algorithmic}[1]
\REQUIRE Patch-wise token sequence $\mathbb{E}_x^{(l)}:(B,W,D)$
\ENSURE Output features for the next encoder layer $\mathbb{E}_x^{(l+1)}:(B,W,D)$
\FOR{$dir$ in $\{forward,backward\}$}
\IF{$dir=backward$}
\STATE $\mathbb{E}_{x,dir}^{(l)}:(B,W,D)\leftarrow\mathrm{Flip}(\mathbb{E}_x^{(l)},\mathrm{dim}=1)$
\ELSE
\STATE $\mathbb{E}_{x,dir}^{(l)}:(B,W,D)\leftarrow\mathbb{E}_x^{(l)}$
\ENDIF
\STATE $\tilde{\mathbb{E}}_{y,dir}^{(l)}:(B,W,D)\leftarrow\mathrm{TSMamba}(\mathbb{E}_{x,dir}^{(l)})$
\STATE $\tilde{\mathbb{E}}_{y,dir}^{(l)}:(B,W,D)\leftarrow\mathrm{Add\&Norm}(\tilde{\mathbb{E}}_{y,dir}^{(l)},\mathbb{E}_{x,dir}^{(l)})$
\ENDFOR
\STATE $\tilde{\mathbb{E}}_y^{(l+1)}:(B,W,D)\leftarrow\mathbb{E}_{y,forward}^{(l)}+\mathbb{E}_{y,backward}^{(l)}$
\STATE $\tilde{\mathbb{E}}_x^{(l+1)}:(B,W,D)\leftarrow\mathrm{FeedForward}(\tilde{\mathbb{E}}_y^{(l+1)})$
\STATE $\mathbb{E}_x^{(l+1)}:(B,W,D)\leftarrow\mathrm{Add\&Norm}(\tilde{\mathbb{E}}_x^{(l+1)},\tilde{\mathbb{E}}_{y}^{(l+1)})$
\STATE \textbf{return} $\mathbb{E}_x^{(l+1)}$
\end{algorithmic}
\label{alg1}
\end{algorithm}

\begin{figure}[!htbp]
    \centering
    \subfigure[Bi-Mamba+ encoder]{
    \includegraphics[width=0.47\linewidth]{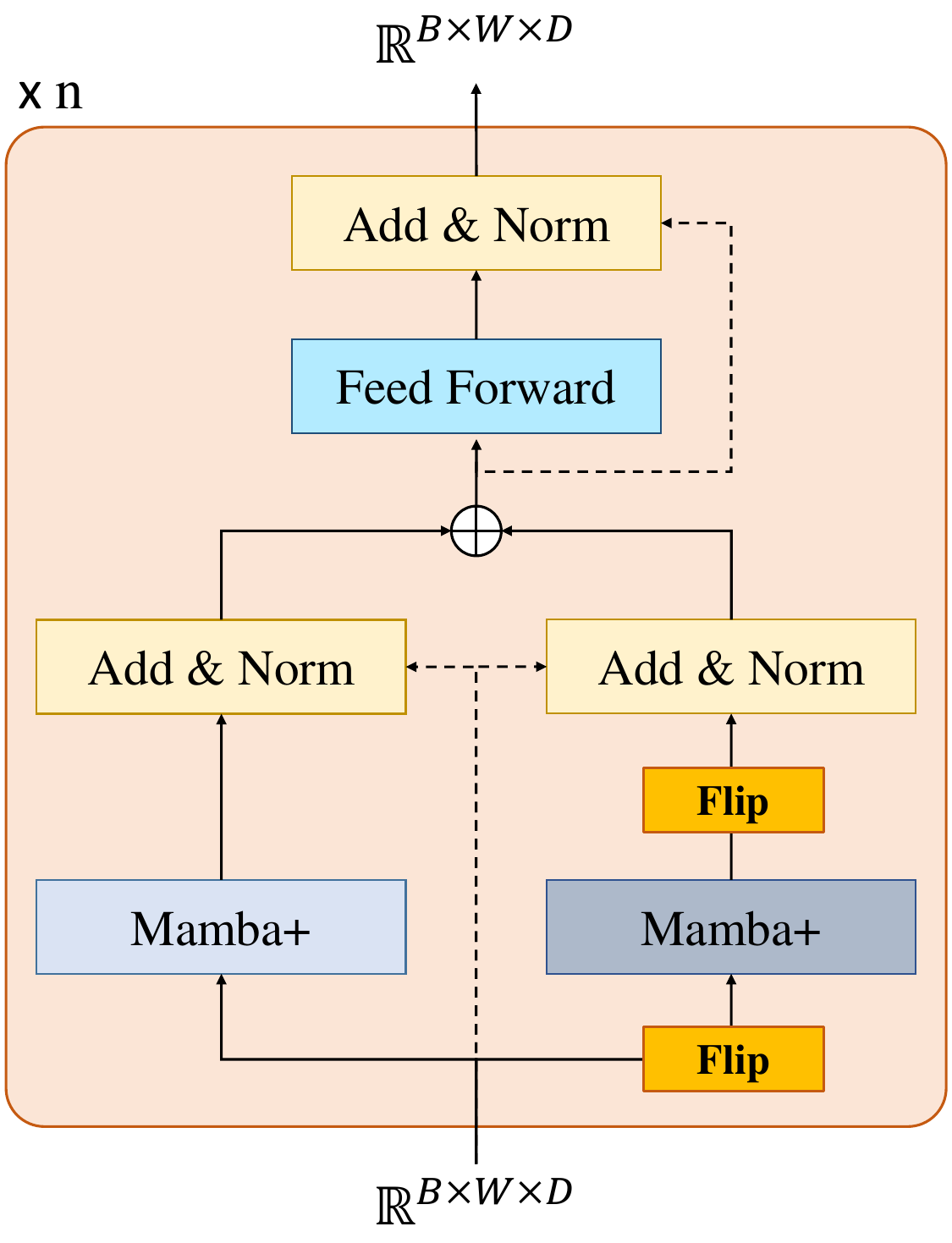}
  }
  \subfigure[Mamba+ block]{
    \includegraphics[width=0.47\linewidth]{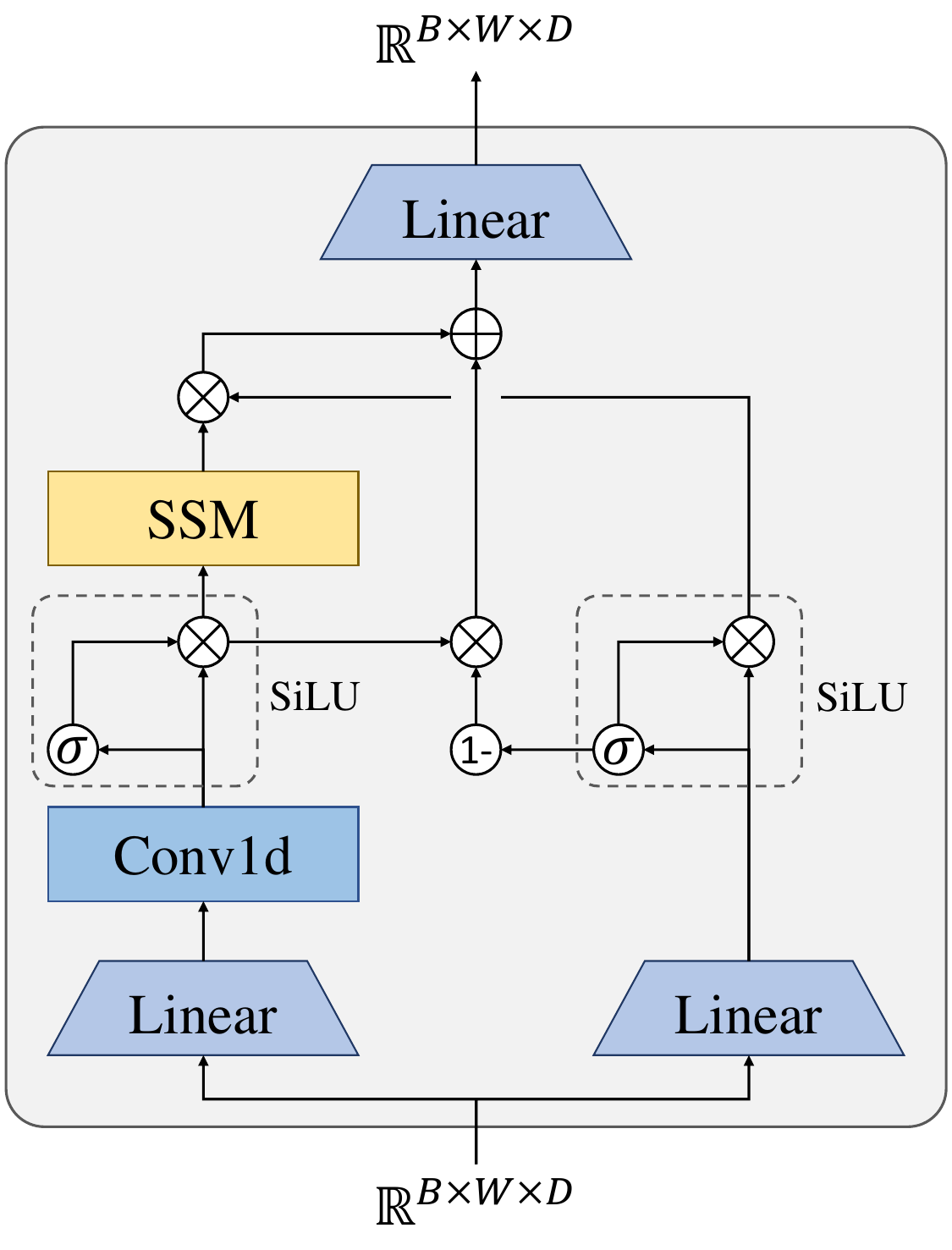}
    }
    \caption{The architecture of (a) Bi-Mamba+ encoder and (b) Mamba+ block.}
    \label{fig:mamba block}
\end{figure}

\subsection{Bidirectional Mamba+ Encoder}
An original Mamba block is designed to process 1-D sequence on one direction. Considering the rich evolutionary patterns of time series in different directions as well as the complexity of inter-series dependencies, we design a bidirectional Mamba+ structure to comprehensively model the MTS data.

A Bi-Mamba+ encoder takes $\mathbb{E}_x^{(l)}\in\mathbb{R}^{B\times W\times D}$ as input, where $l$ refers to the encoder layer, $B$ and $W$ corresponds to $M$ or $J$ depending on the tokenization strategy. Specifically, if $ts=1$, $\mathbb{E}_x^{(l)}\in\mathbb{R}^{J\times M\times D}$ and $\mathbb{E}_x^{(0)}=\mathbb{E}_{mix}$, otherwise, $\mathbb{E}_x^{(l)}\in\mathbb{R}^{M\times J\times D}$ and $\mathbb{E}_x^{(0)}=\mathbb{E}_{ind}$.
There are two Mamba+ blocks in one Bi-Mamba+ encoder to model the input sequence from the forward and backward directions respectively, as shown in Fig \ref{fig:mamba block}(a). We denote the input of these two directions as $\mathbb{E}_{x,dir}^{(l)}$ where $dir\in\{forward,backward\}$. 
We get $\mathbb{E}_x^{(l+1)}=\sum_{dir}^{\{forward,backward\}}\mathcal{F}(\mathbb{E}_{y,dir}^{(l)},\mathbb{E}_{x,dir}^{(l)})$ as the input of the next Bi-Mamba+ encoder layer. The function $\mathcal{F}$ here represents the combination of residual connection and feed forward neural network layers like those in a Transformer encoder. $\mathbb{E}_{y,forward}^{(l)}$ and $\mathbb{E}_{y,backward}^{(l)}$ are the outputs of the two Mamba+ blocks with two modeling directions respectively.
The detailed implementation is shown in Alg.\ref{alg:alg3}.

\subsection{Loss Function}
We use Mean Squared Error (MSE) Loss, which measures the average squared difference between the predicted values and the ground truth. The definition of the Loss function is as follows:
\begin{equation}
    \mathcal{L}(Y,\hat{Y}) = \frac{1}{|Y|}\sum_{i=1}^{|Y|}(y_{(i)}-\hat{y}_{(i)})^2
\label{eq:mae loss}
\end{equation}
where $\hat{Y}$ is the predicted values and $Y$ is the ground truth.

\section{Experiments}
\subsection{Datasets}
We choose 8 real-world datasets to evaluate our proposed model: Weather, Traffic, Electricity, Solar
and 4 ETT datasets(ETTh1, ETTh2, ETTm1, ETTm2). 
These datasets are widely used, covering multiple fields including weather, transportation and energy management. The statistics of the datasets are shown in Table \ref{tab:data}.

\begin{table}[h]
    \centering
    \caption{Statistics of all datasets.}
    \label{tab:data}
    \scalebox{0.9}{
        \begin{tabular}{llll}
        \toprule 
        Dataset & Variables & Frequency & Length \\
        \midrule  
        Weather     & 21  & 10 min & 52696 \\
        Electricity & 321 & 1 hour & 26304 \\
        Traffic     & 862 & 1 hour & 17544 \\
        ETTh1       & 7   & 1 hour & 17420 \\
        ETTh2       & 7   & 1 hour & 17420 \\
        ETTm1       & 7   & 15 min & 69680 \\
        ETTm2       & 7   & 15 min & 69680 \\
        Solar       & 137 & 10min  & 52179\\
        \bottomrule 
        \end{tabular}
    }
\end{table}
\begin{table}[h]
    \centering
    \renewcommand{\arraystretch}{1.1}
    \caption{The information of the experimental platform.}
    \scalebox{0.9}{
    \begin{tabular}{l|l}
    \toprule
    Component & Description \\
    \hline
    System & Ubuntu 20.04 LTS Linux \\
    \hline
    CPU model & Intel(R) Xeon(R) CPU E5-2686 v4 \\
    \hline
    Memory Size & 64GB \\
    \hline
    GPU model & NVIDIA Tesla V100 32GB $(\times 2)$ \\
    \hline
    CUDA Version & 11.8 \\
    \hline
    Python Version & 3.10.14 \\
    \hline
    Pytorch Version & 2.1.1 \\
    \bottomrule
    \end{tabular}
    }
    \label{tab:platform}
\end{table}

\begin{table*}[t]
\centering
\caption{Experimental results of long-term multivariate time series forecasting task on 8 real-world datasets. The best results are in \textbf{bold} and the second best results are \uline{underlined}.}
\label{tab:result total}
\renewcommand{\arraystretch}{1.2}
\resizebox{1.0\linewidth}{!}{
    \begin{tabular}{cc|c|cc|cc|cc|cc|cc|cc|cc|cc|cc|ccc}
        \cline{2-23}
        &\multicolumn{2}{c|}{Models}& \multicolumn{2}{c|}{Bi-Mamba+}& \multicolumn{2}{c|}{S-Mamba}& \multicolumn{2}{c|}{iTransformer}& \multicolumn{2}{c|}{PatchTST}& \multicolumn{2}{c|}{Crossformer}& \multicolumn{2}{c|}{Autoformer}& \multicolumn{2}{c|}{DLinear} & \multicolumn{2}{c|}{TimesNet} & \multicolumn{2}{c|}{CrossGNN}&
        \multicolumn{2}{c}{WITRAN}\\
        \cline{2-23}
        &\multicolumn{2}{c|}{Metric}&MSE&MAE&MSE&MAE&MSE&MAE&MSE&MAE&MSE&MAE&MSE&MAE&MSE&MAE&MSE&MAE&MSE&MAE&MSE&MAE\\
        \cline{2-23}
        &\multirow{4}*{\rotatebox{90}{Weather}}& 96    
        & \uline{0.159} & \textbf{0.205} & 0.166 & \uline{0.210} & 0.174 & 0.214 & 0.178 & 0.219 & \textbf{0.158} & 0.230 & 0.266 & 0.336 & 0.196 & 0.235 & 0.172 & 0.220 & 0.186 & 0.237 & 0.178 & 0.223 \\
        &\multicolumn{1}{c|}{}& 192   
        & \textbf{0.205} & \textbf{0.249} & 0.215 & \uline{0.253} & 0.221 & 0.254 & 0.224 & 0.259 & \uline{0.206} & 0.277 & 0.307 & 0.367 & 0.241 & 0.271 & 0.219 & 0.261 & 0.233 & 0.273 & 0.223 & 0.261 \\
        &\multicolumn{1}{c|}{}& 336   
        & \textbf{0.264} & \textbf{0.291} & 0.276 & \uline{0.298} & 0.278 & \uline{0.298} & 0.292 & 0.306 & \uline{0.272} & 0.335 & 0.359 & 0.395 & 0.292 & 0.306 & 0.280 & 0.306 & 0.289 & 0.312 & 0.288 & 0.309 \\
        &\multicolumn{1}{c|}{}& 720   
        & \textbf{0.343} & \textbf{0.344} & \uline{0.353} & 0.349 & 0.358 & 0.349 & 0.354 & \uline{0.348} & 0.398 & 0.418 & 0.419 & 0.428 & 0.363 & 0.353 & 0.365 & 0.359 & 0.356 & 0.352 & 0.372 & 0.363 \\
        \cline{2-23}
        &\multirow{4}*{\rotatebox{90}{Traffic}}& 96    
        & \textbf{0.375} & \textbf{0.258} & \uline{0.381} & \uline{0.261} & 0.395 & 0.268 & 0.457 & 0.295 & 0.522 & 0.290 & 0.613 & 0.388 & 0.647 & 0.384 & 0.593 & 0.321 & 0.676 & 0.407 & 1.037 & 0.441 \\
        &\multicolumn{1}{c|}{} & 192   
        & \textbf{0.394} & \uline{0.269} & \uline{0.397} & \textbf{0.267} & 0.417 & 0.276 & 0.471 & 0.299 & 0.530 & 0.293 & 0.616 & 0.382 & 0.596 & 0.359 & 0.617 & 0.336 & 0.631 & 0.386 & 1.061 & 0.455 \\
        &\multicolumn{1}{c|}{}& 336   
        & \textbf{0.406} & \textbf{0.274} & \uline{0.423} & \uline{0.276} & 0.433 & 0.283 & 0.482 & 0.304 & 0.558 & 0.305 & 0.616 & 0.382 & 0.601 & 0.361 & 0.629 & 0.336 & 0.640 & 0.387 & 1.095 & 0.470 \\
        &\multicolumn{1}{c|}{}& 720   
        & \textbf{0.440} & \textbf{0.288} & \uline{0.458} & \uline{0.300} & 0.467 & 0.302 & 0.514 & 0.322 & 0.589 & 0.328 & 0.622 & 0.337 & 0.642 & 0.381 & 0.640 & 0.350 & 0.681 & 0.402 & 1.121 & 0.474 \\
        \cline{2-23}
        &\multirow{4}*{\rotatebox{90}{Electricity}}& 96    
        & \textbf{0.140} & \textbf{0.238} & \uline{0.142} & \textbf{0.238} & 0.148 & 0.240 & 0.174 & 0.259 & 0.219 & 0.314 & 0.201 & 0.317 & 0.206 & 0.288 & 0.168 & 0.272 & 0.217 & 0.304 & 0.237 & 0.335 \\
        &\multicolumn{1}{c|}{}& 192   
        & \textbf{0.155} & \textbf{0.253} & 0.169 & 0.267 & \uline{0.162} & \textbf{0.253} & 0.178 & 0.265 & 0.231 & 0.322 & 0.222 & 0.334 & 0.206 & 0.290 & 0.184 & 0.289 & 0.216 & 0.306 & 0.258 & 0.350 \\
        &\multicolumn{1}{c|}{}& 336   
        & \textbf{0.170} & \textbf{0.269} & \uline{0.178} & 0.275 & \uline{0.178} & \textbf{0.269} & 0.196 & 0.282 & 0.246 & 0.337 & 0.231 & 0.338 & 0.220 & 0.305 & 0.198 & 0.300 & 0.232 & 0.321 & 0.273 & 0.362 \\
        &\multicolumn{1}{c|}{}& 720   
        & \textbf{0.197} & \textbf{0.293} & \uline{0.207} & \uline{0.303} & 0.225 & 0.317 & 0.237 & 0.316 & 0.280 & 0.363 & 0.254 & 0.361 & 0.252 & 0.337 & 0.220 & 0.320 & 0.273 & 0.352 & 0.300 & 0.382 \\
        \cline{2-23}
        &\multirow{4}*{\rotatebox{90}{ETTh1}}& 96    
        & \textbf{0.378} & \textbf{0.395} & 0.386 & 0.406 & 0.386 & 0.405 & 0.393 & 0.408 & 0.423 & 0.448 & 0.449 & 0.459 & 0.387 & 0.406 & 0.384 & 0.402 & \uline{0.382} & \uline{0.398} & 0.414 & 0.419 \\
        &\multicolumn{1}{c|}{}& 192   
        & \textbf{0.427} & \uline{0.428} & 0.448 & 0.444 & 0.441 & 0.436 & 0.445 & 0.434 & 0.471 & 0.474 & 0.500 & 0.482 & 0.439 & 0.435 & 0.436 & 0.429 & \textbf{0.427} & \textbf{0.425} & 0.464 & 0.448 \\
        &\multicolumn{1}{c|}{}& 336   
        & \textbf{0.471} & \textbf{0.445} & 0.494 & 0.468 & 0.487 & 0.458 & \uline{0.474} & \uline{0.451} & 0.570 & 0.546 & 0.521 & 0.496 & 0.493 & 0.457 & 0.491 & 0.469 & 0.486 & 0.487 & 0.516 & 0.478 \\
        &\multicolumn{1}{c|}{}& 720   
        & \textbf{0.470} & \textbf{0.457} & 0.493 & 0.488 & 0.503 & 0.491 & \uline{0.480} & \uline{0.471} & 0.653 & 0.621 & 0.514 & 0.512 & 0.490 & 0.478 & 0.521 & 0.500 & 0.482 & 0.477 & 0.538 & 0.509 \\
        \cline{2-23}
        &\multirow{4}*{\rotatebox{90}{ETTh2}}& 96    
        & \textbf{0.291} & \textbf{0.342} & 0.298 & \uline{0.349} & \uline{0.297} & \uline{0.349} & 0.302 & 0.348 & 0.745 & 0.584 & 0.346 & 0.388 & 0.305 & 0.352 & 0.340 & 0.374 & 0.302 & 0.349 & 0.325 & 0.364 \\
        &\multicolumn{1}{c|}{}& 192   
        & \textbf{0.368} & \textbf{0.392} & \uline{0.379} & \uline{0.398} & 0.380 & 0.400 & 0.388 & 0.400 & 0.877 & 0.656 & 0.456 & 0.452 & 0.424 & 0.439 & 0.402 & 0.414 & 0.382 & 0.400 & 0.433 & 0.427 \\
        &\multicolumn{1}{c|}{}& 336   
        & \textbf{0.407} & \textbf{0.424} & \uline{0.417} & \uline{0.432} & 0.428 & \uline{0.432} & 0.426 & 0.433 & 1.043 & 0.731 & 0.482 & 0.486 & 0.456 & 0.473 & 0.452 & 0.452 & 0.421 & 0.439 & 0.471 & 0.457 \\
        &\multicolumn{1}{c|}{}& 720   
        & \textbf{0.421} & \uline{0.439} & 0.431 & 0.449 & \uline{0.428} & \textbf{0.432} & 0.431 & 0.446 & 1.104 & 0.763 & 0.515 & 0.511 & 0.476 & 0.493 & 0.462 & 0.468 & 0.437 & 0.458 & 0.499 & 0.480 \\
        \cline{2-23}
        &\multirow{4}*{\rotatebox{90}{ETTm1}}& 96    
        & \textbf{0.320} & \textbf{0.360} & 0.331 & 0.368 & 0.334 & 0.368 & \uline{0.329} & \uline{0.367} & 0.404 & 0.426 & 0.505 & 0.475 & 0.353 & 0.374 & 0.338 & 0.375 & 0.340 & 0.374 & 0.375 & 0.402 \\
        &\multicolumn{1}{c|}{}& 192   
        & \textbf{0.361} & \textbf{0.383} & 0.371 & 0.387 & 0.377 & 0.391 & \uline{0.367} & \uline{0.385} & 0.450 & 0.451 & 0.553 & 0.496 & 0.389 & 0.391 & 0.374 & 0.387 & 0.377 & 0.390 & 0.427 & 0.434 \\
        &\multicolumn{1}{c|}{}& 336   
        & \textbf{0.386} & \textbf{0.402} & 0.417 & 0.418 & 0.426 & 0.420 & \uline{0.399} & 0.419 & 0.532 & 0.515 & 0.621 & 0.537 & 0.421 & 0.413 & 0.410 & 0.411 & 0.401 & \uline{0.407} & 0.455 & 0.452 \\
        &\multicolumn{1}{c|}{}& 720   
        & \textbf{0.445} & \textbf{0.437} & 0.471 & 0.448 & 0.491 & 0.459 & 0.454 & \uline{0.439} & 0.666 & 0.589 & 0.671 & 0.561 & 0.484 & 0.448 & 0.478 & 0.450 & \uline{0.453} & 0.442 & 0.527 & 0.488 \\
        \cline{2-23}
        &\multirow{4}*{\rotatebox{90}{ETTm2}} & 96    
        & \uline{0.176} & 0.263 & 0.179 & 0.263 & 0.180 & 0.264 & \textbf{0.175} & \textbf{0.259} & 0.287 & 0.366 & 0.255 & 0.339 & 0.182 & 0.264 & 0.187 & 0.267 & 0.177 & \uline{0.261} & 0.191 & 0.272 \\
        &\multicolumn{1}{c|}{}& 192   
        & 0.242 & 0.304 & 0.253 & 0.310 & 0.250 & 0.309 & \uline{0.241} & \uline{0.302} & 0.414 & 0.492 & 0.281 & 0.340 & 0.257 & 0.315 & 0.249 & 0.309 & \textbf{0.240} & \textbf{0.298} & 0.261 & 0.316 \\
        &\multicolumn{1}{c|}{}& 336   
        & \textbf{0.304} & \uline{0.344} & 0.312 & 0.348 & 0.311 & 0.348 & \uline{0.305} & \textbf{0.343} & 0.597 & 0.542 & 0.339 & 0.372 & 0.318 & 0.353 & 0.321 & 0.351 & \uline{0.305} & 0.345 & 0.330 & 0.358 \\
        &\multicolumn{1}{c|}{}& 720   
        & \textbf{0.402} & 0.402 & 0.412 & 0.408 & 0.412 & 0.407 & \textbf{0.402} & \textbf{0.400} & 1.730 & 1.042 & 0.433 & 0.432 & 0.426 & 0.419 & 0.408 & 0.403 & 0.403 & \textbf{0.400} & 0.450 & 0.427 \\
        \cline{2-23}
        &\multirow{4}*{\rotatebox{90}{Solar}}& 96    
        & \textbf{0.184} & \textbf{0.222} & 0.205 & 0.241 & \uline{0.203} & \uline{0.237} & 0.234 & 0.286 & 0.310 & 0.331 & 0.884 & 0.711 & 0.290 & 0.378 & 0.250 & 0.292 & 0.222 & 0.314 & 0.238 & 0.285 \\
        &\multicolumn{1}{c|}{}& 192   
        & \textbf{0.225} & \textbf{0.254} & 0.237 & 0.270 & \uline{0.233} & \uline{0.261} & 0.267 & 0.310 & 0.734 & 0.725 & 0.834 & 0.692 & 0.320 & 0.398 & 0.296 & 0.318 & 0.240 & 0.318 & 0.282 & 0.302 \\
        &\multicolumn{1}{c|}{}& 336   
        & \uline{0.249} & \textbf{0.270} & 0.256 & 0.286 & \textbf{0.248} & \uline{0.273} & 0.290 & 0.315 & 0.750 & 0.735 & 0.941 & 0.723 & 0.353 & 0.415 & 0.319 & 0.330 & 0.264 & 0.324 & 0.329 & 0.345 \\
        &\multicolumn{1}{c|}{}& 720   
        & \uline{0.250} & \textbf{0.273} & 0.260 & 0.288 & \textbf{0.249} & \uline{0.275} & 0.289 & 0.317 & 0.769 & 0.765 & 0.882 & 0.717 & 0.356 & 0.413 & 0.338 & 0.337 & 0.260 & 0.321 & 0.326 & 0.348 \\
        \cline{2-23}
    \end{tabular}
}
\end{table*}

\subsection{Baseline Models}

We compare our model with 4 Transformer-based models, 1 MLP-based model, 1 CNN-based model, 1 RNN-based model, 1 GNN-based model and 1 SSM-based model. These models cover the mainstream design for time series forecasting tasks and achieve SOTA performance in their respective model categories. The detailed descriptions are as follows:
\begin{itemize}
    \item \textbf{Autoformer} \citep{autoformer} uses a series decomposition technique with Auto-Correlation mechanism to capture cross-time dependency for long-term time series forecasting.
    \item \textbf{PatchTST} \citep{patchtst} adopts patching and channel independent techniques, making semantic extraction of time series from a single time step to multiple time steps.
    \item \textbf{Crossformer} \citep{crossformer} is aware of the fact that segmenting subsequences in LSTF is beneficial. It uses the same patching strategy as PatchTST. To additinally capture inter-series dependencies, it designs to use another attention layer working with a routing mechanism to reduce complexity.
    \item \textbf{iTransformer} \citep{liu2023itransformer} inverts the modeling method of Transformer by using self-attention to capture inter-series dependencies. 
    \item \textbf{DLinear} \citep{dlinear} decomposes time series into two different components, and generates a single Linear layer for each of them. Such a simple design has defeated all the complex transformer models proposed before it.
    \item \textbf{TimesNet} \citep{wu2022timesnet}  extends the analysis of temporal variations into the 2-D space by transforming the 1-D time series into a set of 2-D tensors based on multiple periods.
    \item \textbf{WITRAN} \citep{jia2024witran} designs a novel RNN structure that process the univariate input sequence in the 2-D space with a fixed scale.
    \item \textbf{CrossGNN} \citep{huang2024crossgnn} is a model mainly based on GNN. It views time series in a multi-scale way. It uses GNN to capture both cross-scale and cross-series dependencies.
    \item \textbf{S-Mamba} \citep{wang2024mamba} generates embeddings for each time series through a simple MLP layer and uses Mamba to extract inter-series dependencies.
\end{itemize}

\subsection{Experimental Settings}
\begin{figure*}[!htbp]
    \centering
    \subfigure[Weather]{
    \includegraphics[width=0.22\linewidth]{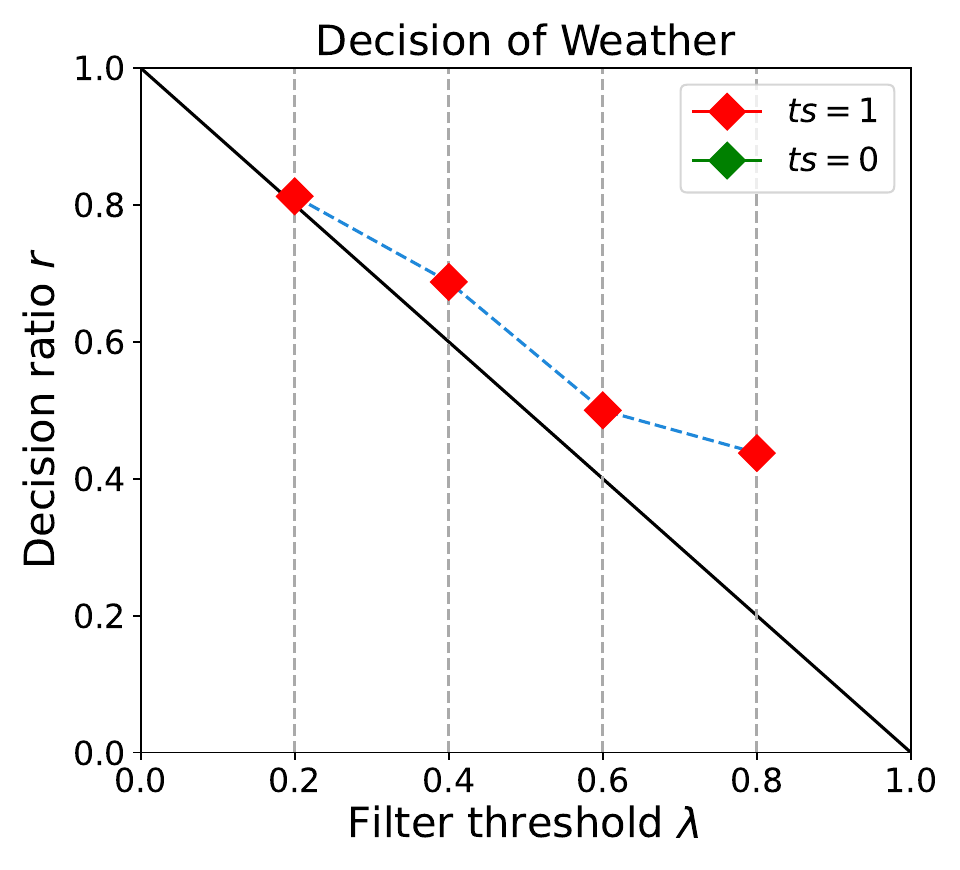}
    }
    \subfigure[Traffic]{
    \includegraphics[width=0.22\linewidth]{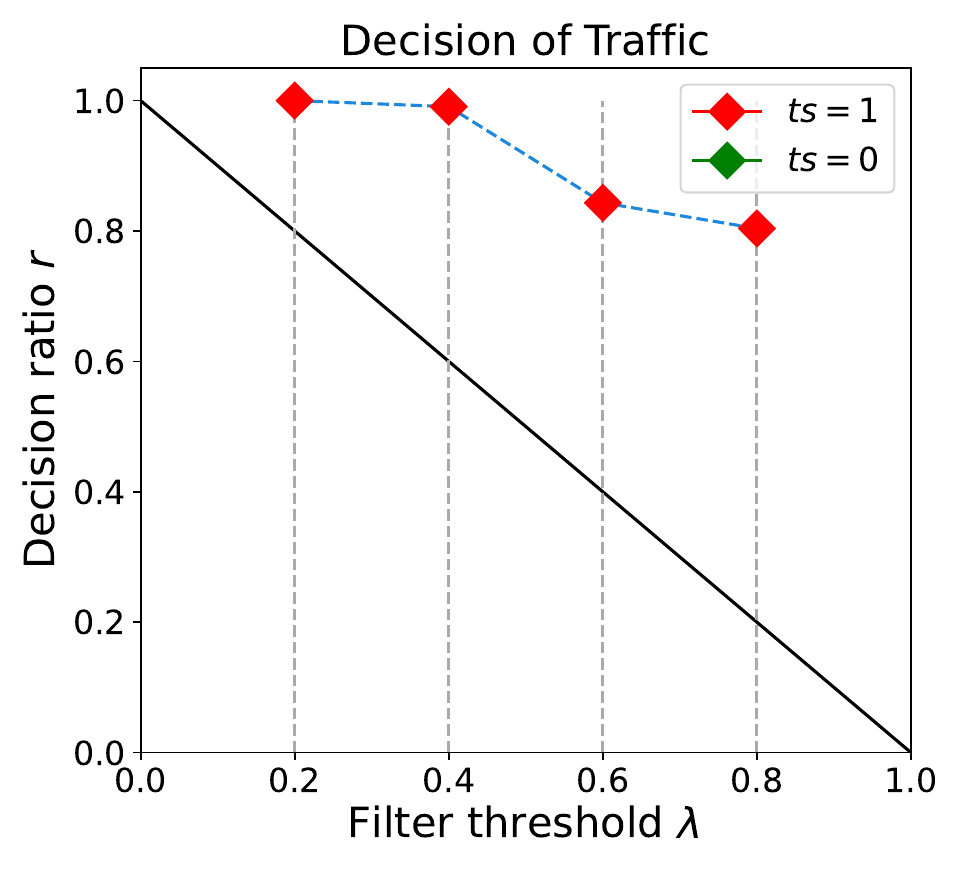}
    }
    \subfigure[Electricity]{
    \includegraphics[width=0.22\linewidth]{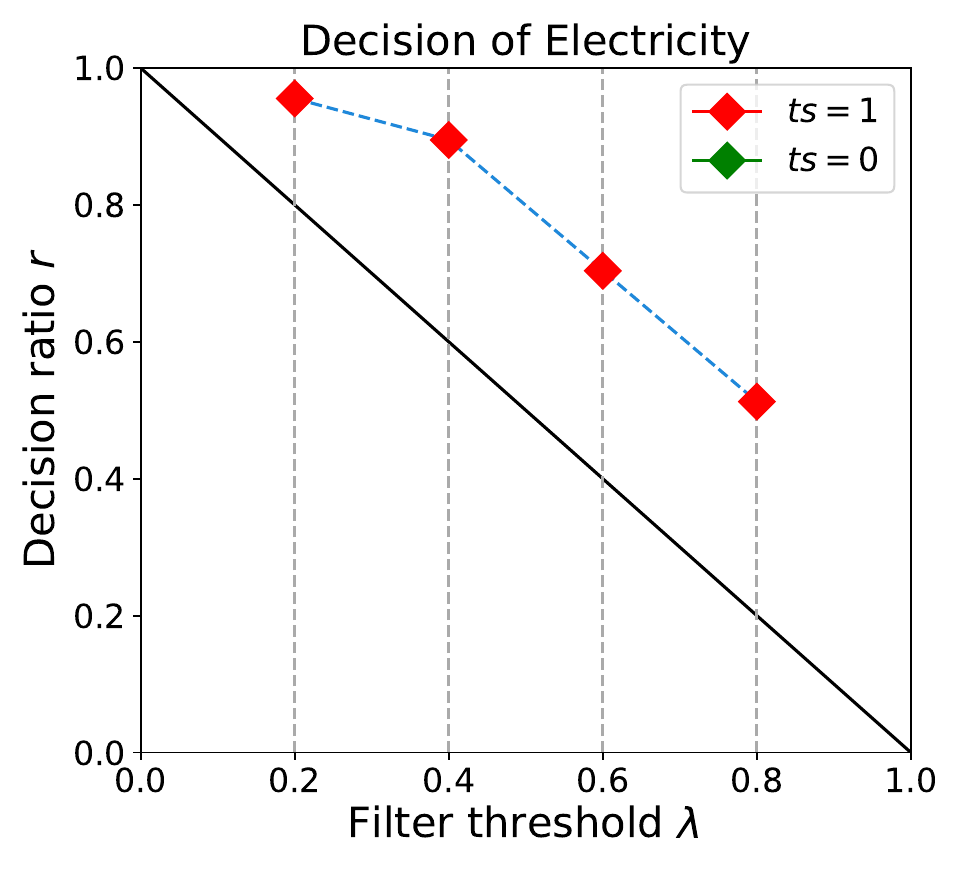}
    }
    \subfigure[Solar]{
    \includegraphics[width=0.22\linewidth]{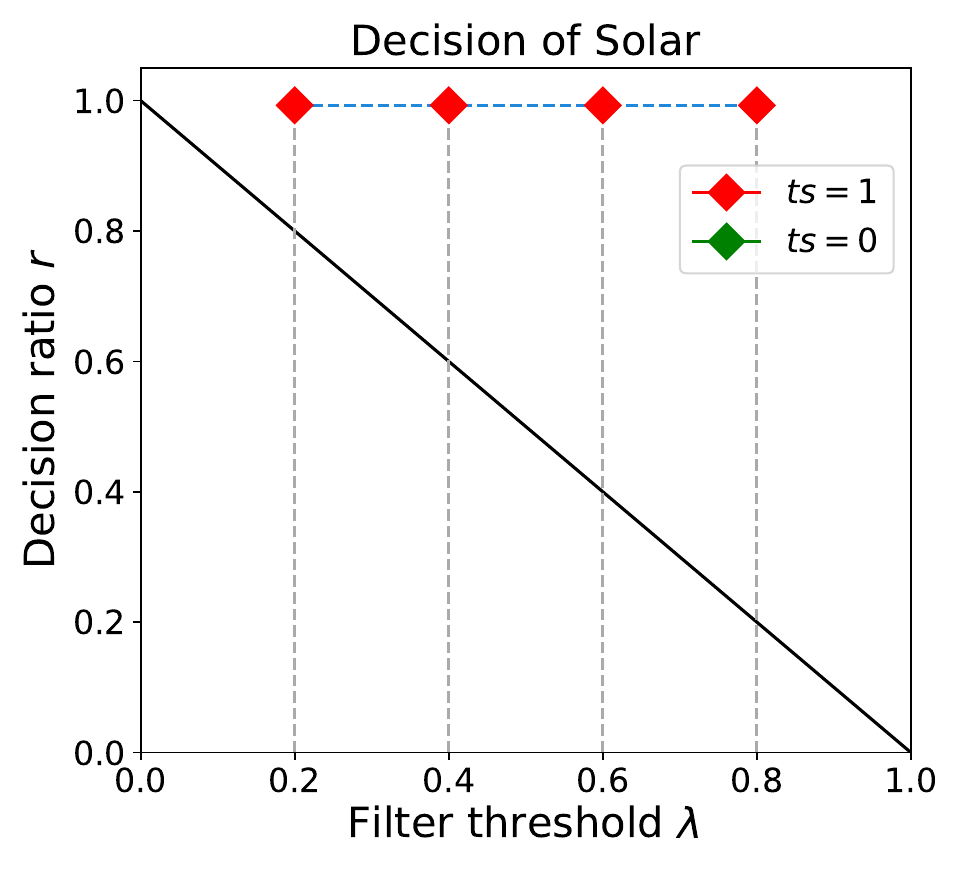}
    }
    \subfigure[ETTh1]{
    \includegraphics[width=0.22\linewidth]{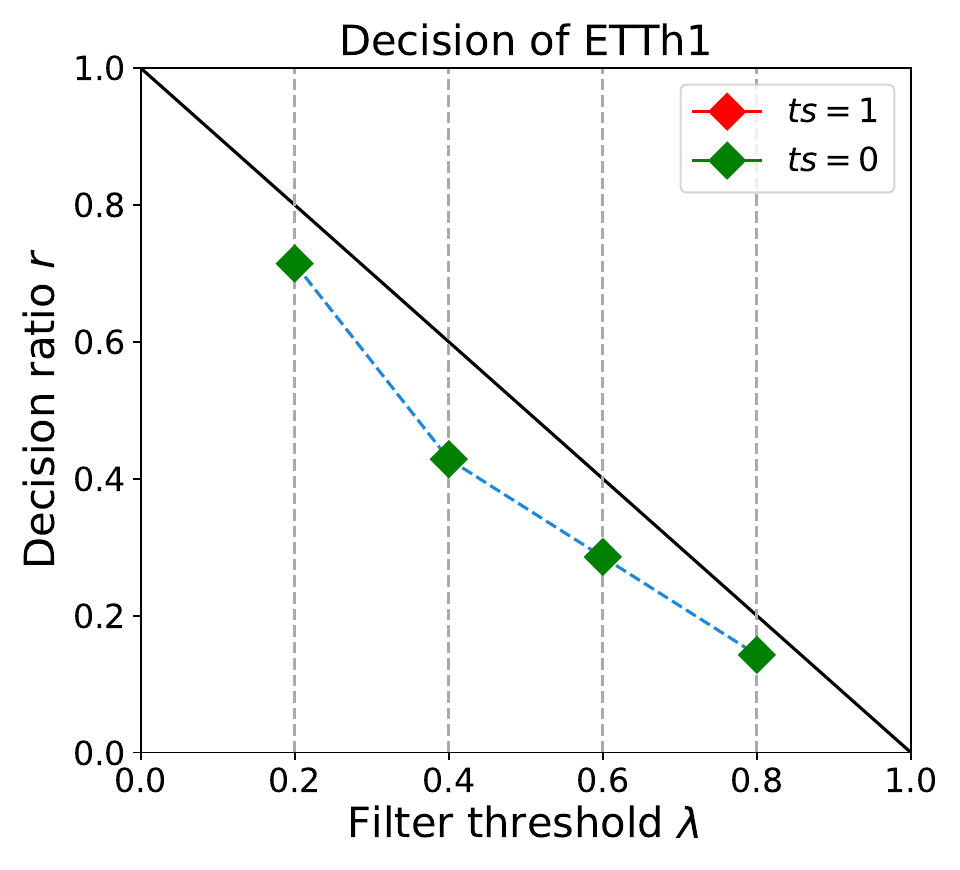}
    }
    \subfigure[ETTh2]{
    \includegraphics[width=0.22\linewidth]{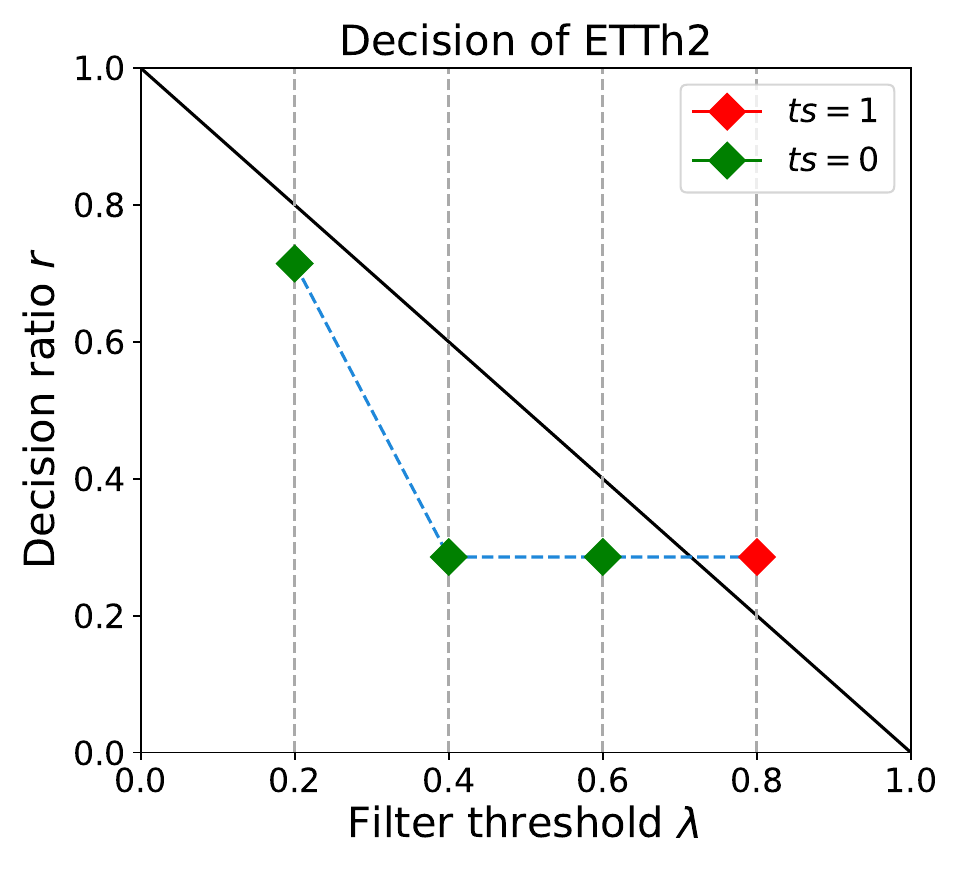}
    }
    \subfigure[ETTm1]{
    \includegraphics[width=0.22\linewidth]{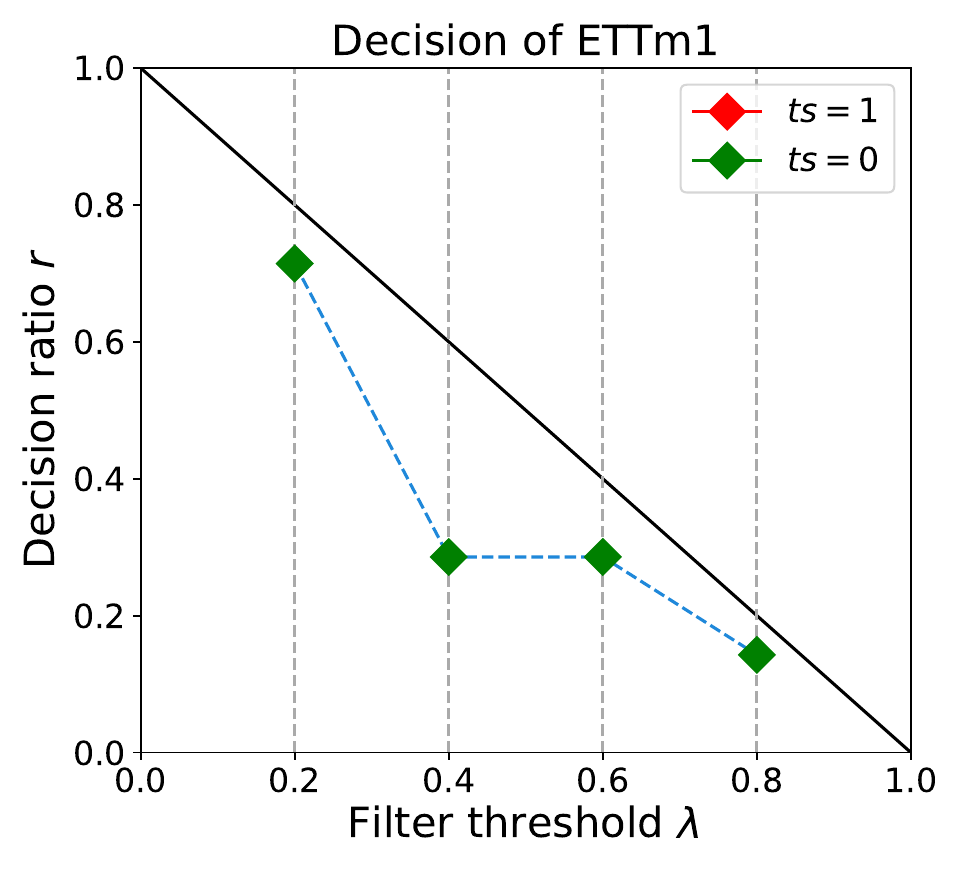}
    }
    \subfigure[ETTm2]{
    \includegraphics[width=0.22\linewidth]{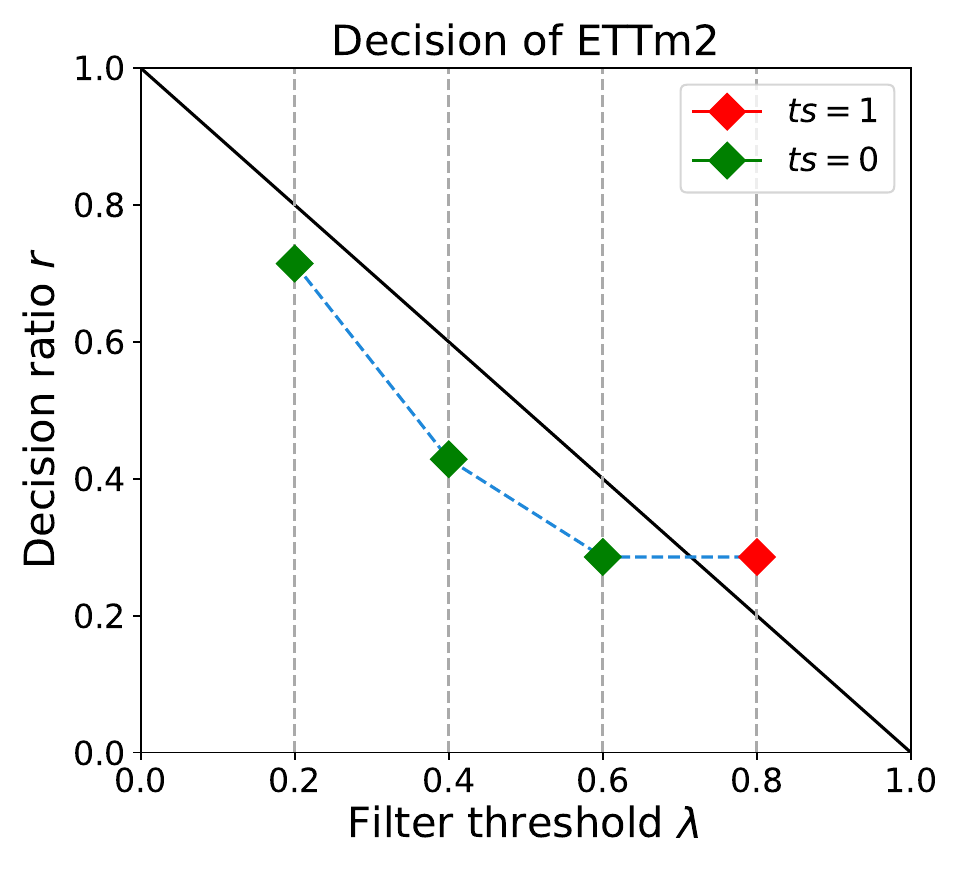}
    }
    \caption{The tokenization strategy indicator $ts$ of different datasets with varying settings of filter threshold $\lambda$.}
    \label{fig:decision maker}
\end{figure*}

We set the same historical look-back window size of $L=96$ for all models on all datasets. The predicting length is set to $H\in\{96,192,336,720\}$. We set $S=\frac{1}{2}P$ and use patch length $P=\frac{1}{4}L$ by default. Considering that different models may have different sensitivity of granularity of local information, we maintain the patching settings consistent with the original papers for PatchTST and Crossformer. 
For the SRA decider, we set $\lambda=0.6$ because generally when the Spearman coefficient correlation is larger than $0.6$, it can be inferred that the two sequences have a strong positive correlation. More detailed evaluation of $\lambda$ is shown in Fig. \ref{fig:decision maker}.
Bi-Mamba+ splits a sequence into patches, which mainly retain local information but contains less global information compared to a whole sequence. Considering that datasets with more variables may contain more complex evolutionary patterns, for Bi-Mamba+, PatchTST and Crossformer that use patching technique, we set $D=128$ for Weather, Traffic, Electricity, Solar and $D=64$ for ETT datasets, while for S-Mamba and iTransformer that map the whole sequence to tokens, we set $D=512$ for Weather, Traffic, Electricity, Solar and $D=256$ for ETT datasets.
As for parameters within Mamba+ block, we follow \citep{ahamed2024timemachine,wang2024mamba} and set convolutional kernel size $d\_conv=2$ and hidden state expansion $expand=1$ on all datasets. We set hidden dimension $d\_state=16$ for Weather, Electricity and Traffic and $d\_state=8$ for ETT datasets.
We evaluate Bi-Mamba+ with encoder layer $l\in\{1,2,3\}$ and conduct grid search for learning rate on [5e-5, 1e-4, 2e-4, 5e-4, 1e-3, 2e-3, 5e-3] to find the best results. For the metrics to evaluate the performance, we utilize MSE and MAE. All models are optimized using the ADAM algorithm\citep{kingma2014adam} and early stopping mechanism. The training process is limited to 40 epochs.

\subsection{Experimental Environment}
To exclude the influence of the model training environment, we conduct all experiments on the same device with 2 NVIDIA Tesla V100 (32G) GPUs. The detailed information of our experimental platform is shown in table \ref{tab:platform}.
\begin{table*}[t]
\centering
\caption{Ablation studies of (a) w/o SRA-I which use channel-independent strategy only, (b) w/o SRA-M which use channel-mixing strategy only, (c) w/o Bi which use forward direction Mamba block only, (d) w/o Residual that removes the residual connection, (e) S-Mamba and (f) PatchTST used for the benchmark models. The best results are in \textbf{bold}.}
\label{tab:ablation}
\renewcommand{\arraystretch}{1.2}
\resizebox{0.6\linewidth}{!}{
\setlength{\tabcolsep}{2.5pt}
    \begin{tabular}{cc|cc|cc|cc|cc|cc|cc|cc}
        \toprule
        \multicolumn{2}{c|}{Models}& \multicolumn{2}{c|}{w/o SRA-I}& \multicolumn{2}{c|}{w/o SRA-M}& \multicolumn{2}{c|}{w/o Bi}& \multicolumn{2}{c|}{w/o Residual} & \multicolumn{2}{c|}{w/o M \& w/ A} & \multicolumn{2}{c|}{S-Mamba} & \multicolumn{2}{c}{PatchTST}\\
        \cmidrule(lr){1-2}\cmidrule(lr){3-4}\cmidrule(lr){5-6}\cmidrule(lr){7-8}\cmidrule(lr){9-10}\cmidrule(lr){11-12}\cmidrule(lr){13-14}\cmidrule(lr){15-16}
        \multicolumn{2}{c|}{Metric}&MSE&MAE&MSE&MAE&MSE&MAE&MSE&MAE&MSE&MAE&MSE&MAE&MSE&MAE\\
        \midrule
        \multicolumn{2}{c|}{Weather} & 0.252 & 0.277 & \textbf{0.243} & \textbf{0.272} & 0.251 & 0.281 & 0.252 & 0.283 & 0.257 & 0.281 & 0.253 & 0.278 & 0.259 & 0.281 \\
        \midrule
        \multicolumn{2}{c|}{Traffic} & 0.453 & 0.287 & \textbf{0.404} & \textbf{0.272} & 0.435 & 0.289 & 0.439 & 0.306 & 0.417 & 0.281 & 0.415 & 0.276 & 0.481 & 0.305 \\
        \midrule
        \multicolumn{2}{c|}{Electricity} & 0.188 & 0.274 & \textbf{0.166} & \textbf{0.263} & 0.176 & 0.271 & 0.179 & 0.279 & 0.174 & 0.269 & 0.174 & 0.271 & 0.196 & 0.281 \\
        \midrule
        \multicolumn{2}{c|}{ETTh1} & \textbf{0.437} & \textbf{0.431} & 0.447 & 0.437 & 0.443 & 0.437 & 0.456 & 0.446 & 0.446 & 0.445 & 0.465 & 0.457 & 0.448 & 0.441 \\
        \midrule
        \multicolumn{2}{c|}{ETTh2} & \textbf{0.372} & \textbf{0.399} & \textbf{0.372} & 0.402 & 0.376 & 0.406 & 0.382 & 0.407 & 0.374 & 0.402 & 0.381 & 0.407 & 0.387 & 0.407 \\
        \midrule
        \multicolumn{2}{c|}{ETTm1} & \textbf{0.378} & \textbf{0.396} & 0.389 & 0.404 & 0.386 & 0.401 & 0.386 & 0.401 & 0.381 & 0.396 & 0.398 & 0.405 & 0.387 & 0.400 \\
        \midrule
        \multicolumn{2}{c|}{ETTm2} & \textbf{0.281} & 0.328 & 0.284 & 0.332 & 0.288 & 0.335 & 0.291 & 0.339 & 0.287 & 0.334 & 0.289 & 0.333 & \textbf{0.281} & \textbf{0.326} \\
        \midrule
        \multicolumn{2}{c|}{Solar} & 0.236 & 0.263 & \textbf{0.227} & \textbf{0.255} & 0.235 & 0.262 & 0.232 & 0.263 & 0.241 & 0.276 & 0.239 & 0.271 & 0.270 & 0.307 \\
        \bottomrule
    \end{tabular}
}
\end{table*}

\subsection{Main Results}
Table \ref{tab:result total} shows the results of long-term multivariate forecasting. Our model achieves superior performance compared with other baselines on all datasets with different predicting lengths. Compared with the current SOTA Transformer model iTransformer, Bi-Mamba+ reduces the MSE, MAE errors by 4.72\% and 2.60\% on average. The improvement of MSE and MAE comes to 3.76\% and 2.67\% compared with S-Mamba, which is also an SSM-based model using Mamba. 
It is worth noting that Bi-Mamba+ adopts channel-mixing strategy on Weather, Traffic, Electricity and Solar and channel-independent strategy on 4 ETT datasets. For channel-mixing strategy, the patching approach used by Bi-Mamba+ helps the model capture inter-series dependencies at a finer granularity compared with S-Mamba and iTransformer that transform the whole sequence into a token. For channel-independent strategy, the patch-wise tokens fed into Mamba blocks contain more semantic information, enhancing the model's ability for capturing the inner evolutionary process of each univariate time series. Compared with PatchTST that adopt the same patching technique, our proposed Mamba+ block achieves more accurate predictions than self-attention.

\subsection{Ablation Study}
\label{sec:ablation}
We verify the effectiveness of the SRA decider, the bidirectional module and the residual connection. Before we conduct the ablation experiments, we first export the results of the SRA decider to view the strategy choosing for different datasets. 
For Spearman correlation coefficient $\rho$, we typically use $\lambda>0.6$ to evaluate whether two variables have positive correlation strong enough to influence each other. 
The decision results under varying $\lambda$ are shown in Fig. \ref{fig:decision maker}.
For ETT datasets that have few variables, the model tends to adopt channel-independent strategy to emphasis intra-series dependencies, while for Weather, Traffic and Electricity that have large amount of variables, the model tends to adopt channel-mixing strategy to emphasis inter-series dependencies. We remove the SRA decider to form two specific Bi-Mamba+ models: (a) w/o SRA-I which only uses channel-independent tokenization strategy and (b) w/o SRA-M which only uses channel-mixing tokenization strategy. For the bidirectional Mamba+ encoder, we remove the backward direction Mamba+ to evaluate the effectiveness of bidirectional structure. This setting forms the following model: (c) w/o Bi. Besides, we remove the \textbf{Add\&Norm} layer to evaluate the effect of residual connection: (d) w/o Residual. We also replace Mamba+ blocks with the dot-product Attention: (e) w/o M \& w/ A. 

We calculate the average MSE and MAE scores as shown in table \ref{tab:ablation}. 
Without the SRA decider, the model utilizes either channel-mixing or channel independent tokenization strategies. The MSE and MAE increase by 0.95\% and 0.77\% for w/o SRA-M and 3.56\% and 1.45\% for w/o SRA-I respectively. Therefore, the proposed SRA decider makes the optimal choice of channel-independent and channel-mixing strategies and utilizing different tokenization strategies on different datasets produces better predicting performance. 

When the backward Mamba block is removed, the predicting performance of MSE and MAE is deteriorated 3.25\% and 2.44\%  respectively, which indicates that the design of bidirectional module enhances the model to get more comprehensive understanding of the MTS data. 

The residual network enhances the expressive ability and feature utilization of the model by reusing early features. It can also reduce overfitting in the training process. The MSE and MAE of w/o Residual decrease by 4.37\% and 4.05\% respectively, indicating the effectiveness of the residual connection module. Full results of Ablation Study can be found in table \ref{tab:result ablation total} in Appendix.

It is worth noting that replacing Mamba+ with Attention causes performance degradation, indicating that Mamba may be a better choice for time series forecasting.

\begin{figure*}[t]
    \centering
    \includegraphics[width=1\textwidth]{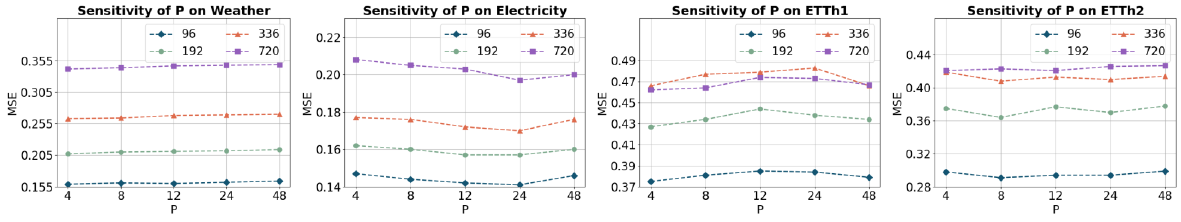}
    \caption{Results of varying patch length $P$ on Weather, Electricity, ETTh1 and ETTh2.}
    \label{fig:hyper2}
\end{figure*}

\subsection{Hyper-parameter Sensitivity Analysis}
\label{sec:hyper}
To verify whether Bi-Mamba+ is sensitive to the hyper-parameters, we conduct experiments on (a) Spearman coefficient filter $\lambda$, (b) length of patches, (c) dropout (d) Conv-1D kernel size of Mamba+ and (e) state dimension size. For each setting, we freeze other parameters that are not currently fine-tuned and repeat the experiment 5 times with 40 epochs (with early-stop control). We record the average MSE of $H\in\{96,192,336,720\}$. Overall, different parameter settings do not cause significant changes in the results, which also shows the robustness of our proposed model.

\begin{figure}[!htbp]
    \centering
    \subfigure[Sensitivity of dropout on\newline Electricity]{
    \includegraphics[width=0.47\linewidth]{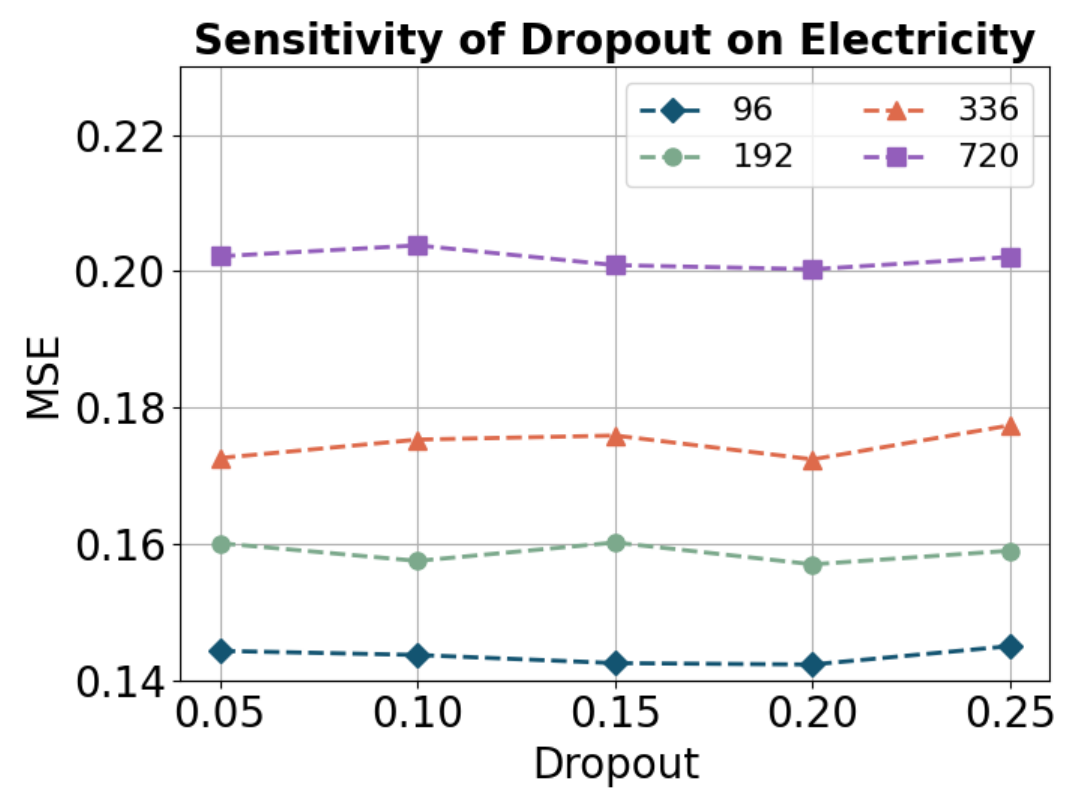}
    }
    \subfigure[Sensitivity of dropout on\newline ETTh2]{
    \includegraphics[width=0.47\linewidth]{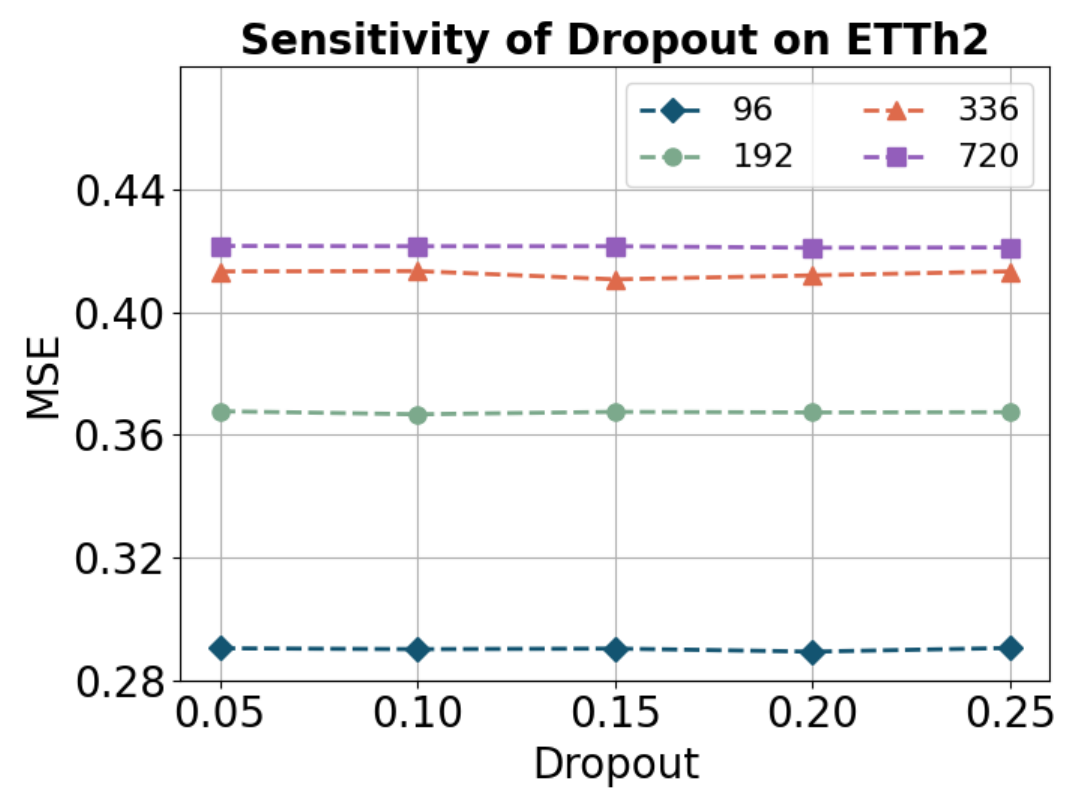}
    }
    \subfigure[Sensitivity of hidden state\newline dimension on Electricity]{
    \includegraphics[width=0.47\linewidth]{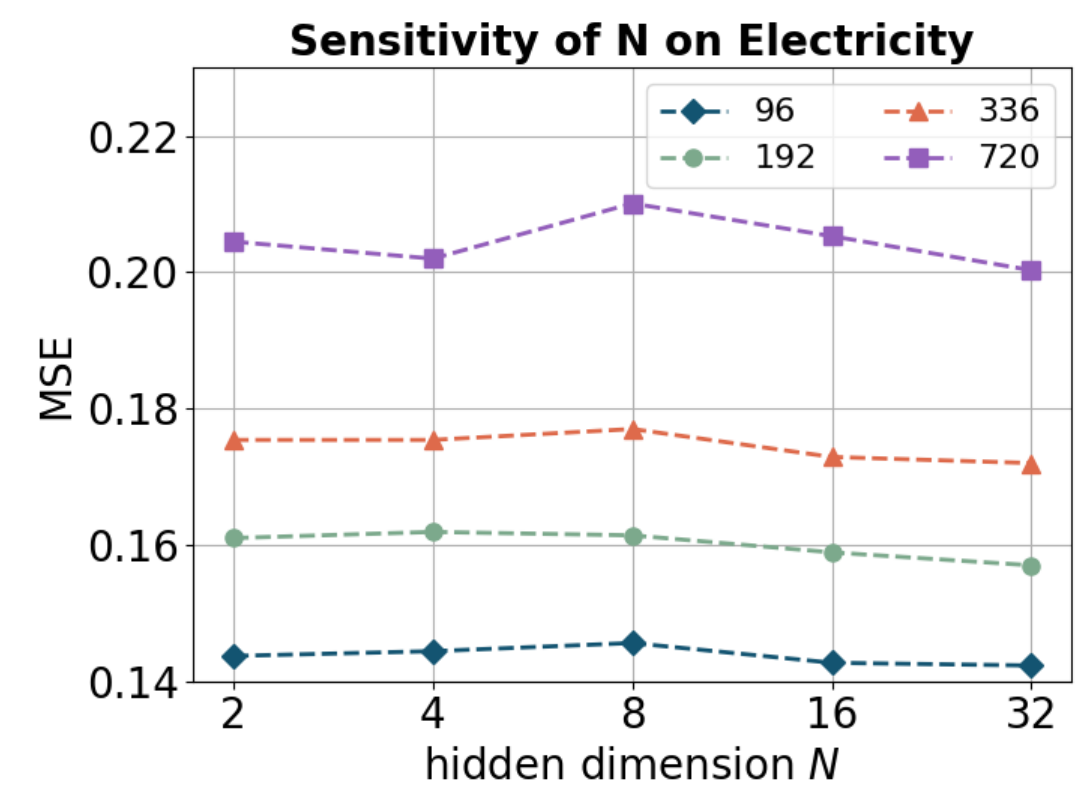}
    }
    \subfigure[Sensitivity of hidden state\newline dimension on ETTh2]{
    \includegraphics[width=0.47\linewidth]{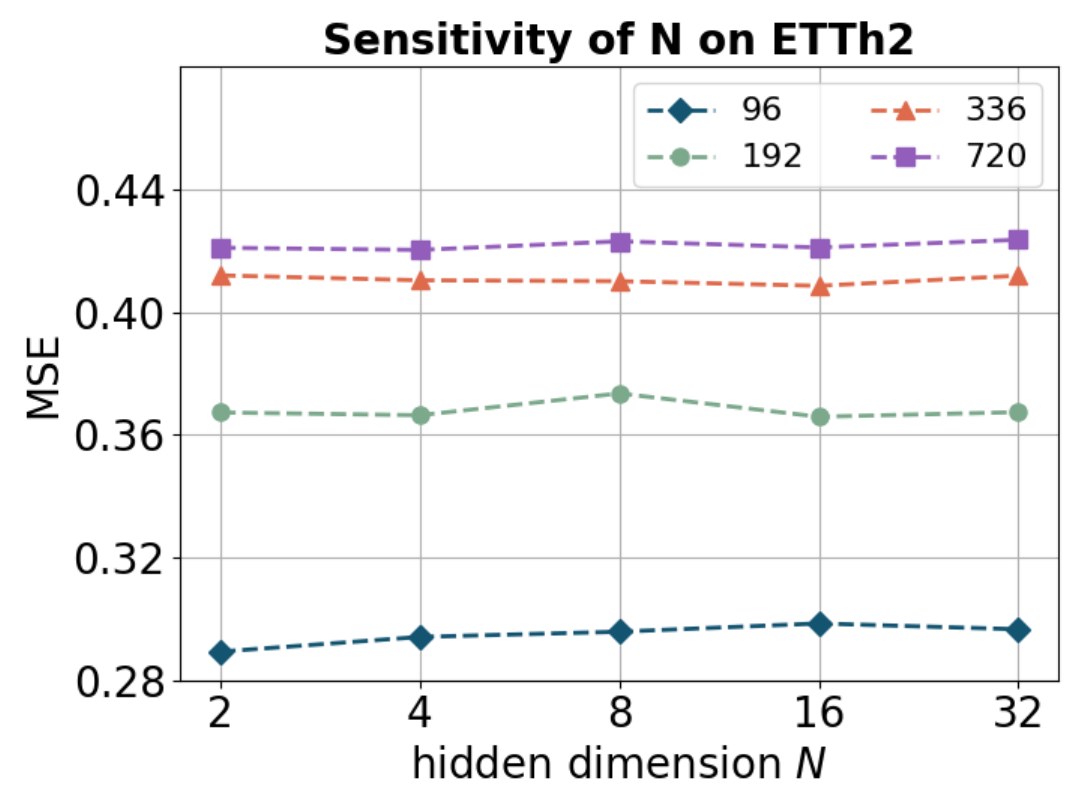}
    }
    \subfigure[Sensitivity of kernel\_size\newline on Electricity]{
    \includegraphics[width=0.47\linewidth]{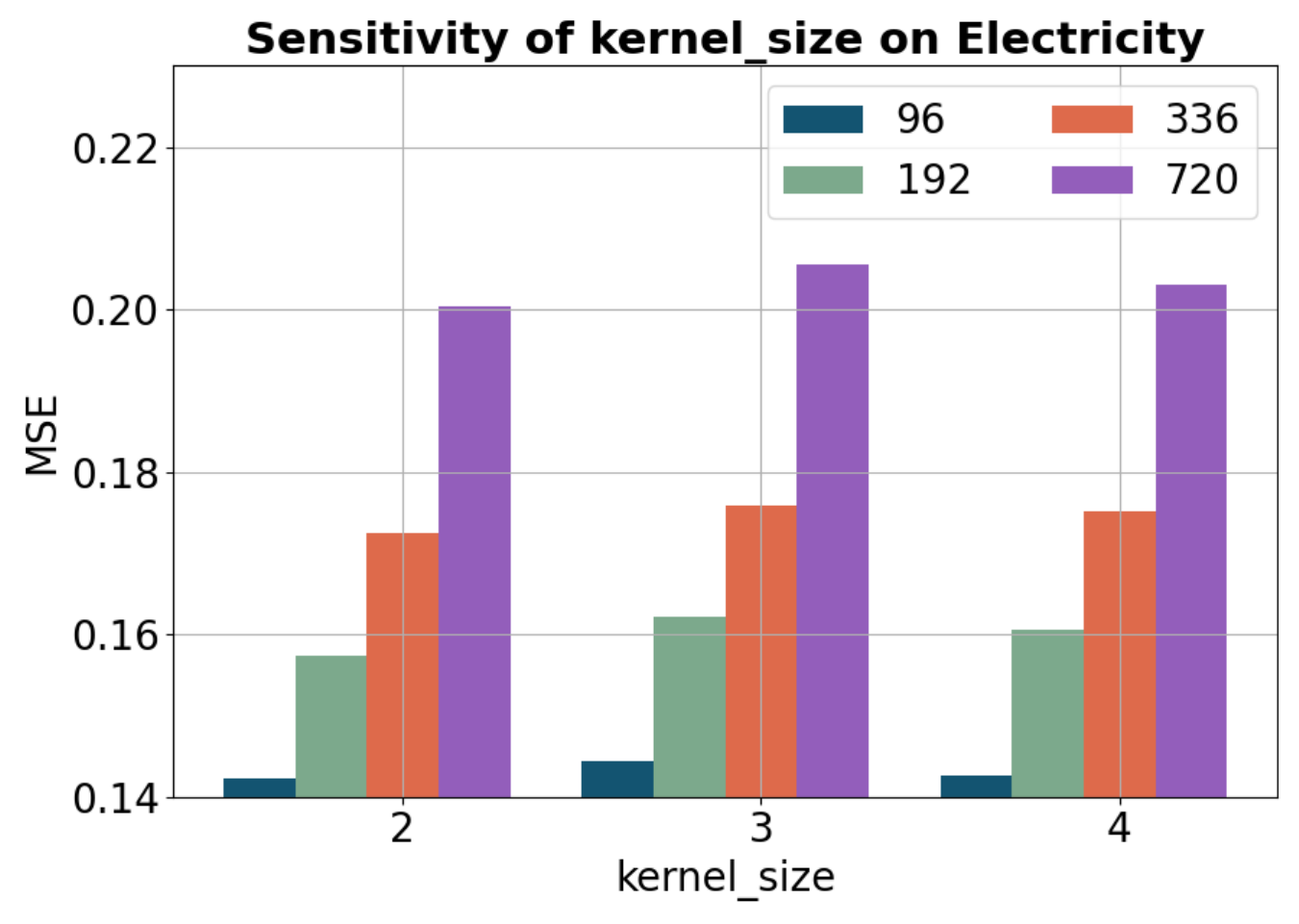}
    }
    \subfigure[Sensitivity of kernel\_size\newline on ETTh2]{
    \includegraphics[width=0.47\linewidth]{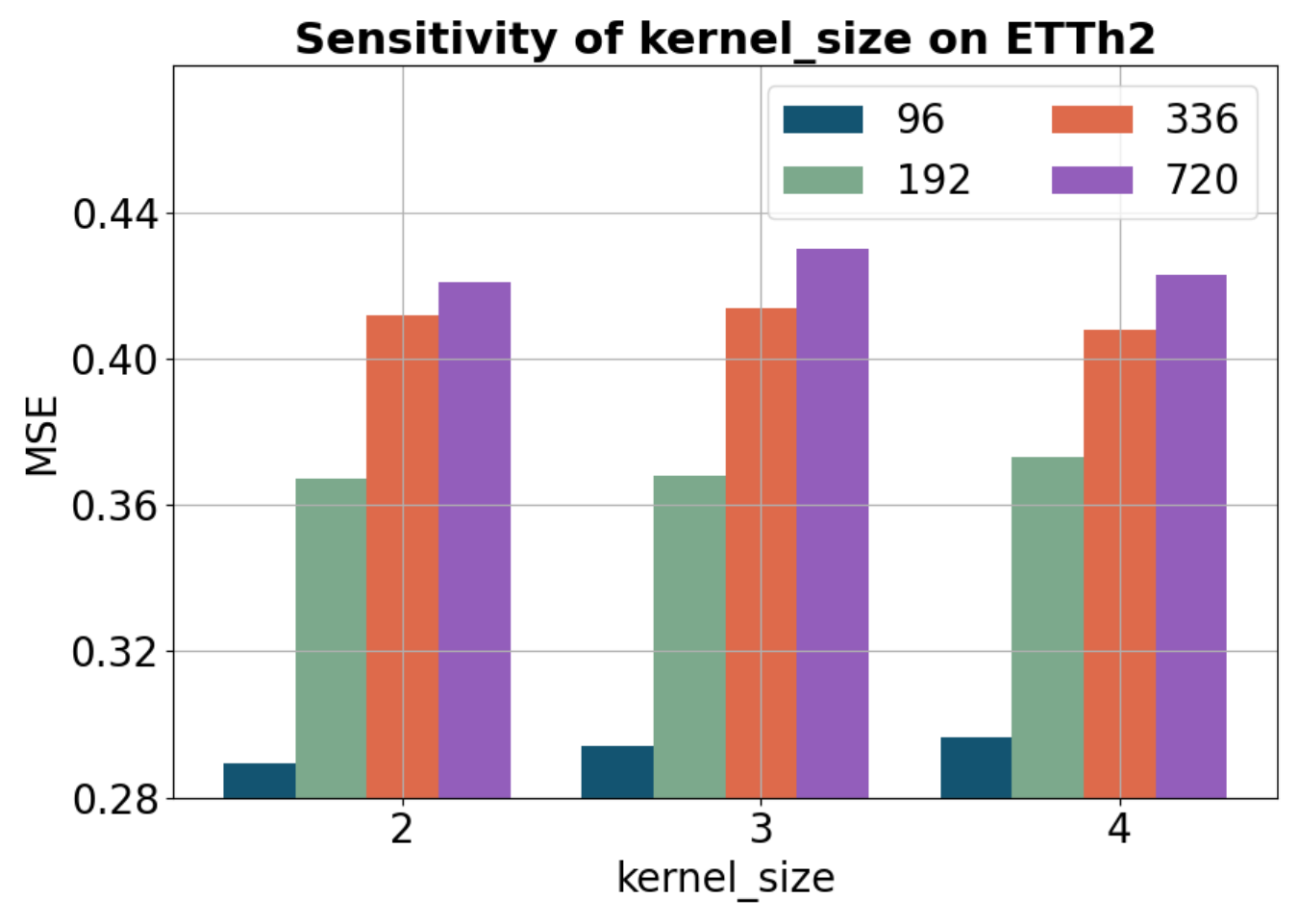}
    }
    \caption{Results of varying dropout, hidden dimension N and kernel size values on Electricity and ETTh2.}
    \label{fig:hyper}
\end{figure}

\subsubsection{Spearman coefficient filter $\lambda$.}
The value $\lambda$ is used for filter out variable pairs that have positive correlations. A larger $\lambda$ corresponds to stronger positive correlations between two sequences. We set $\lambda\in\{0.2,0.4,0.6,0.8\}$ and calculate the decision ratio $r$ and decision indicator $ts$. The results are shown in Fig. \ref{fig:decision maker}. Overall, different $\lambda$ does not frequently cause the change of tokenization strategy. For datasets that have large number of variables, such as Traffic and Electricity, the decision indicator $ts$ prefers to be equal to $1$, while for datasets with less variables such as ETTh1 and ETTh2, $ts$ is more likely to be $0$. Note that the $ts$ may undergo a change from $\lambda=0.6$ to $\lambda=0.8$ on ETTh2 and ETTm2, resulting in different tokenization strategies. The results in table \ref{tab:ablation} indicate some correlations with the phenomenon. For ETTh2 and ETTm2, the tokenization strategy does not notably affect the predicting performance because of the existence of more highly correlated variables.

\subsubsection{Patch length}
We use varying patch length of $P\in\{\frac{1}{2}L,\frac{1}{4}L,\frac{1}{8}L,\frac{1}{12}L,\frac{1}{24}L\}$. We conduct experiments on Weather, Electricity, ETTh1 and ETTh2. For datasets that use channel-independent or channel-mixing tokenization strategy, different patch lengths reflect the intra- or inter-series dependencies at different timescales. The results are shown in Fig. \ref{fig:hyper2}. 
For Weather and ETT datasets, the model gets optimal performance with $P=\frac{1}{24}L$ or $P=\frac{1}{12}L$. These datasets have relatively less variables and the data distribution exhibit stronger non-stationarity locally, hence suitable for more fine-grained modeling. The optimal patch length for Electricity is $P=\frac{1}{4}L$, indicating that coarse-grained modeling is beneficial for processing the data that is overall stable.
In general, varying lengths of patch does not cause significant performance differences, which indicates that our model is robust in capturing intra- and inter-series dependencies at different time scales.

\subsubsection{Dropout}
The Dropout layer randomly masks a portion of the input elements to zero to preventing overfitting. This helps to improve the generalization ability of the model. The dropout layer works in the Add\&Norm layers in Fig \ref{fig:archetecture}. We set $dropout\in\{0.05,0.1,0.15,0.2,0.25\}$. The results can be seen in Fig. \ref{fig:hyper}(a) and Fig. \ref{fig:hyper}(b). In most cases, the optimal result is obtained when $dropout=0.2$, which indicates that a too low $dropout$ value does not work well to improve the generalization ability, while a too large $dropout$ value may cause performance degradation.

\subsubsection{Hidden state dimension of Mamba.}
A higher dimension $N$ of the hidden state can help the model capture more complex system evolutionary patterns, but a too large $N$ will reduce the efficiency of model training and may leads to overfitting. We set $N\in\{2,4,8,16,32\}$ and the results are shown in Fig. \ref{fig:hyper}(c) and Fig. \ref{fig:hyper}(d). For Electricity with more variables, a larger $N$ produces better performance, this indicates that datasets with more complex inter-series dependencies prefer larger hidden state dimension for capturing the evolutionary patterns of time series. For ETTh2 with less variables, different values of $N$ do not leads to varying predicting accuracy, which indicates that the temporal patterns in such datasets is less complex.

\subsubsection{1-D convolutional kernel size of Mamba.}
The 1-D convolution module in Mamba is used for capturing temporal dependencies of the input series. We conduct experiments on different kernel size of 1, 2 and 3. The results in Fig. \ref{fig:hyper}(e) and Fig. \ref{fig:hyper}(f) show that Bi-Mamba+ gets optimal performance with the kernel size of 2. Overall, our model is not sensitive to the kernel size.

\begin{figure}[t]
    \centering
    \includegraphics[width=1.0\linewidth]{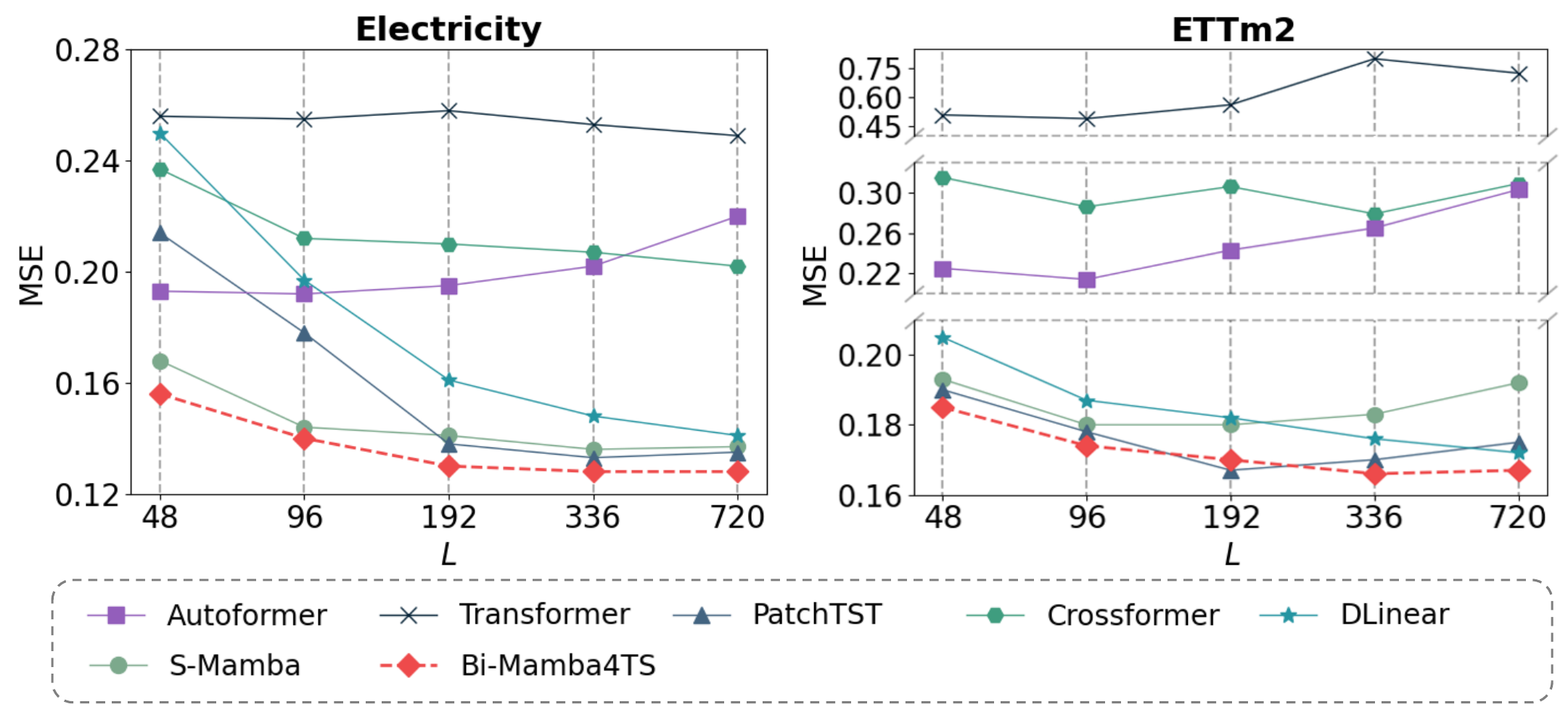}
    \caption{Results of varying look-back window size $L$ on Electricity and ETTm2. The selected baseline models are Autoformer, Transformer, PatchTST, Crossformer, DLinear and S-Mamba. Overall, Bi-Mamba+ gets more accurate prediction with the increased receptive field and shows better predicting performance than other models.}
    \label{fig:L}
\end{figure}

\begin{figure}[t]
    \centering
    \subfigure[ETTh1]{
    \includegraphics[width=1.0\linewidth]{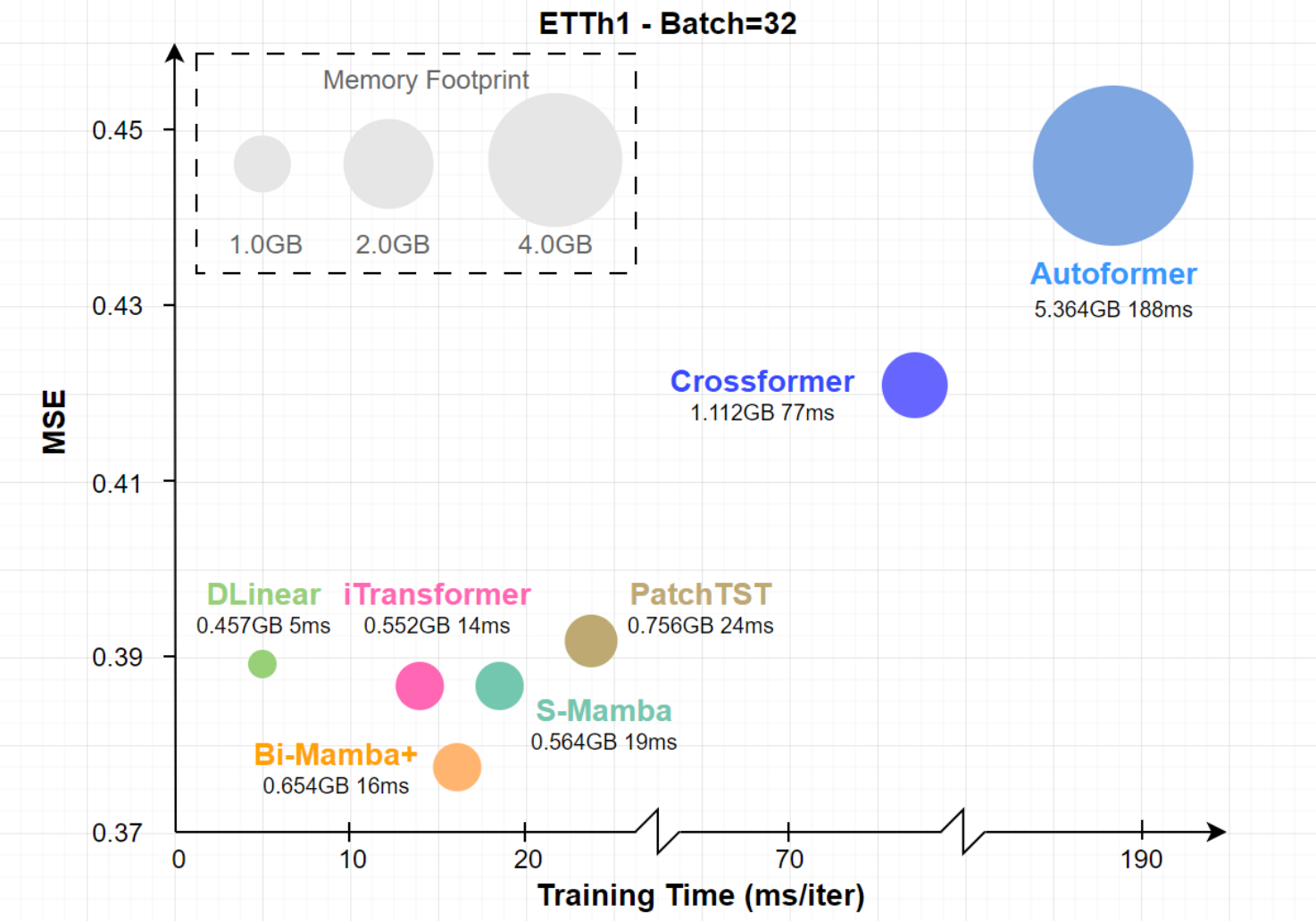}
  }
  \subfigure[Traffic]{
    \includegraphics[width=1.0\linewidth]{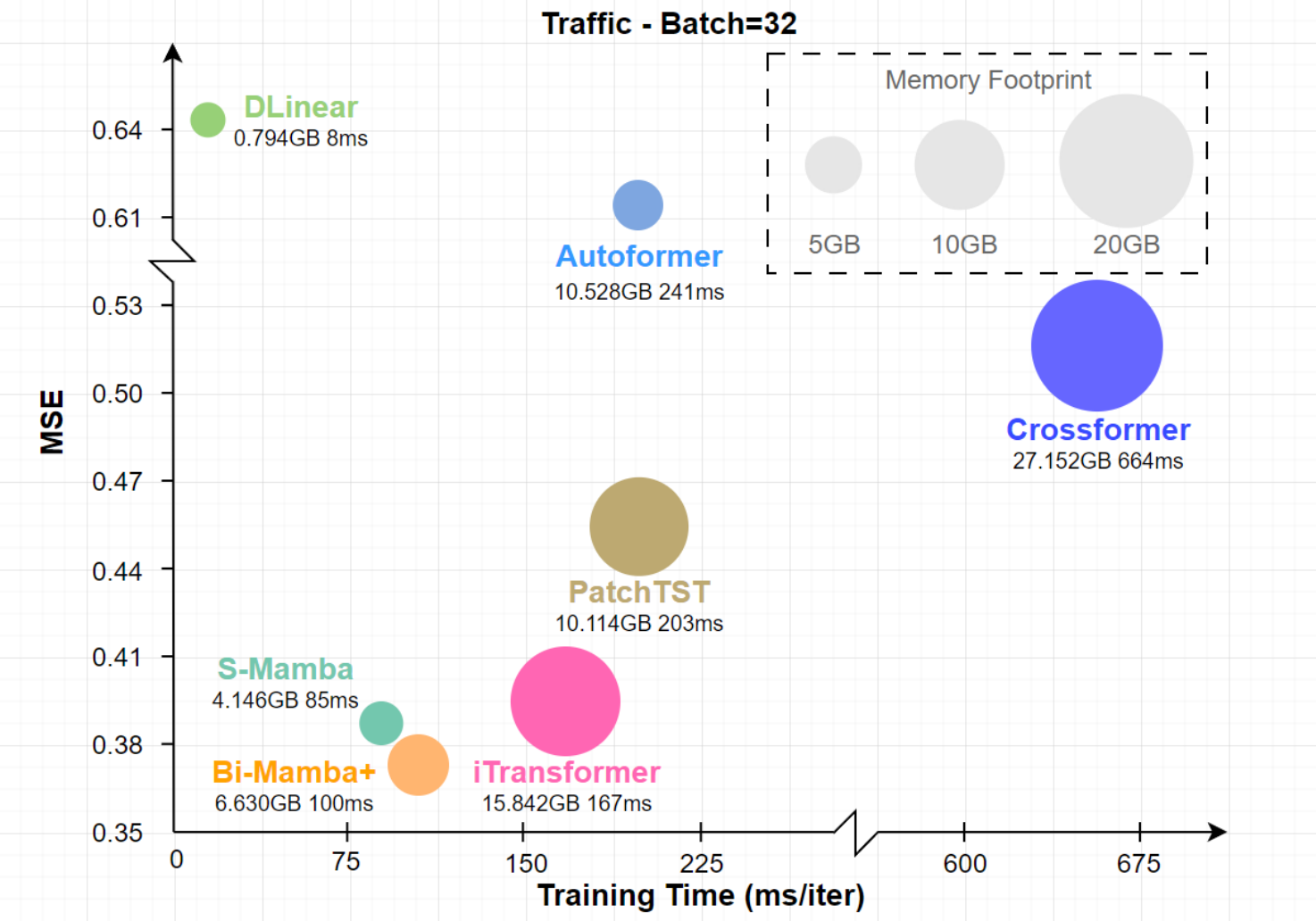}
    }
    \caption{Model efficiency comparison with $L=96,H=96$ on ETTh1 and Traffic. The batch size is set to $32$.}
    \label{fig:efficiency}
\end{figure}

\subsection{Long-term Dependencies Capturing}
To evaluate the long-term dependencies capturing capability of Bi-Mamba+, we set $L\in\{48,96,192,336,720\}$ and record the average MSE scores of Bi-Mamba+, S-Mamba, Transformer, Autoformer, Crossformer and DLinear. We conduct experiments on Electricity and ETTm2. According to the SRA decider, these two datasets show preference to channel-mixing and channel-independent tokenization respectively, thus can more comprehensively reflect the adaptability of different models to the varying temporal evolutionary patterns.

Fig. \ref{fig:L} shows the results of the performance of models with varying look-back window size. Bi-Mamba+, S-Mamba, PatchTST and Crossformer get better predicting performance as the receptive field increases on Electricity. For previous channel-mixing models, Transformer does not show significant changes in predicting performance regardless of the size of the look-back window. Autoformer, however, even faces performance degradation at inflection point $L=96$. These two models may encounter the challenge of overfitting caused by point-wise tokens.
We note that S-Mamba, PatchTST and S-Mamba do not achieve performance improvements similar to those on Electricity as the receptive fields increase on ETTm2. 
Compared to Electricity, the periodicity of ETTm2 is less obvious, making it more difficult to model the evolutionary patterns within the series.
Previous channel-mixing models such as Autoformer encounters performance degradation at inflection point $L=96$, which may related to overfitting caused by point-wise tokens. Overall, Bi-Mamba+ can benefit from longer look-back window size and adapt to more complex temporal evolutionary patterns.

\begin{figure*}[t]
    \centering
    \subfigure[Results of Bi-Mamba+ and iTransformer on Electricity]{
    \includegraphics[width=0.45\linewidth]{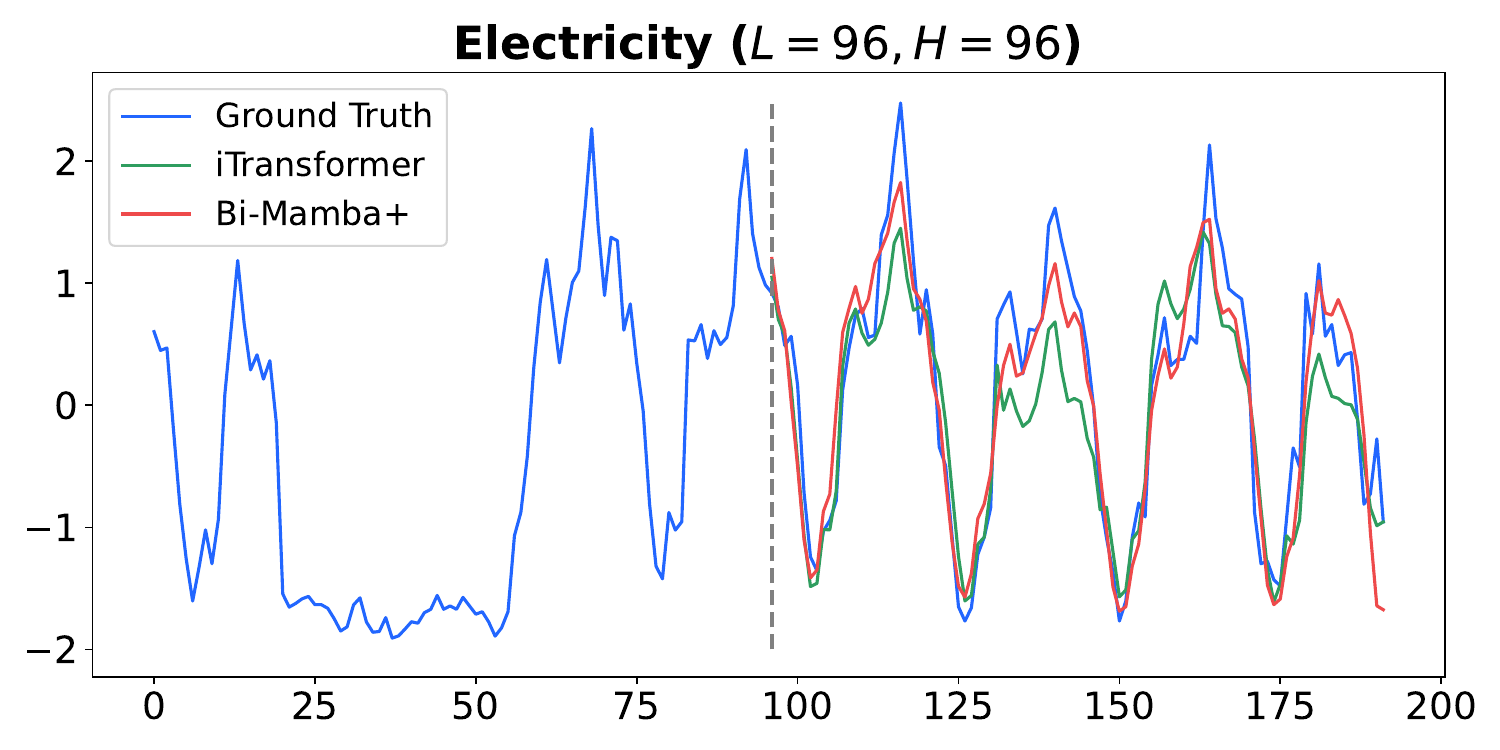}
    }
    \subfigure[Results of Autoformer, DLinear and PatchTST on Electricity]{
    \includegraphics[width=0.45\linewidth]{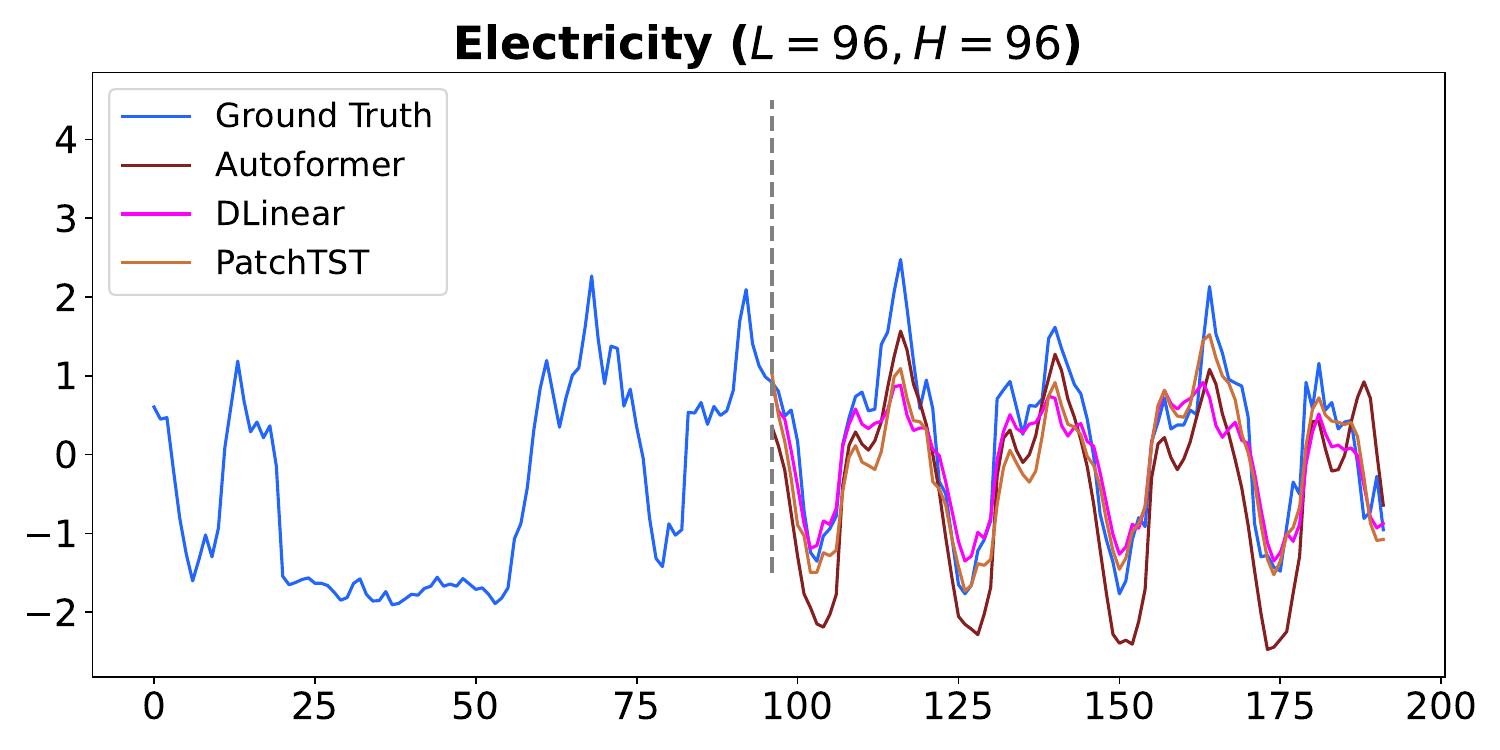}
    }
    \subfigure[Results of Bi-Mamba+ and iTransformer on Traffic]{
    \includegraphics[width=0.45\linewidth]{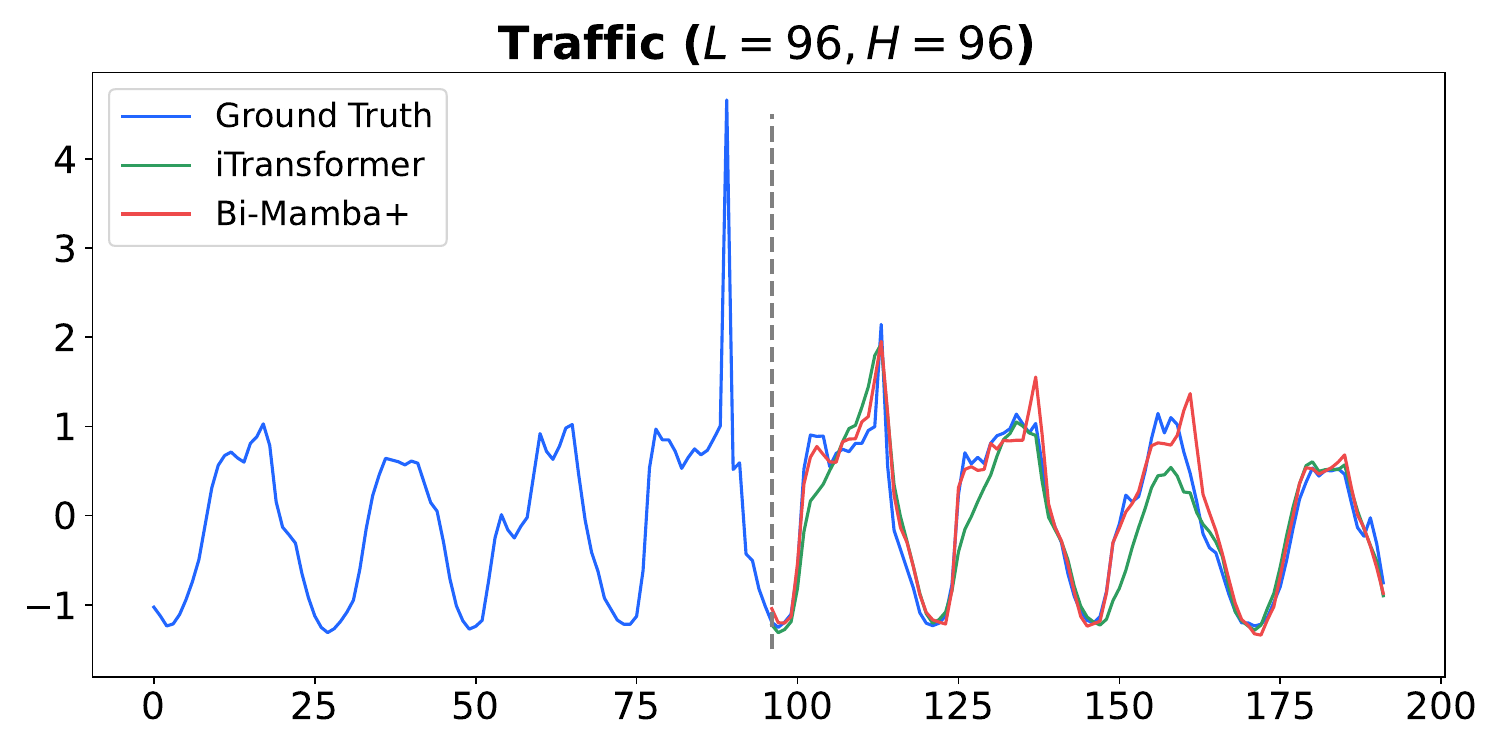}
    }
    \subfigure[Results of Autoformer, DLinear and PatchTST on Traffic]{
    \includegraphics[width=0.45\linewidth]{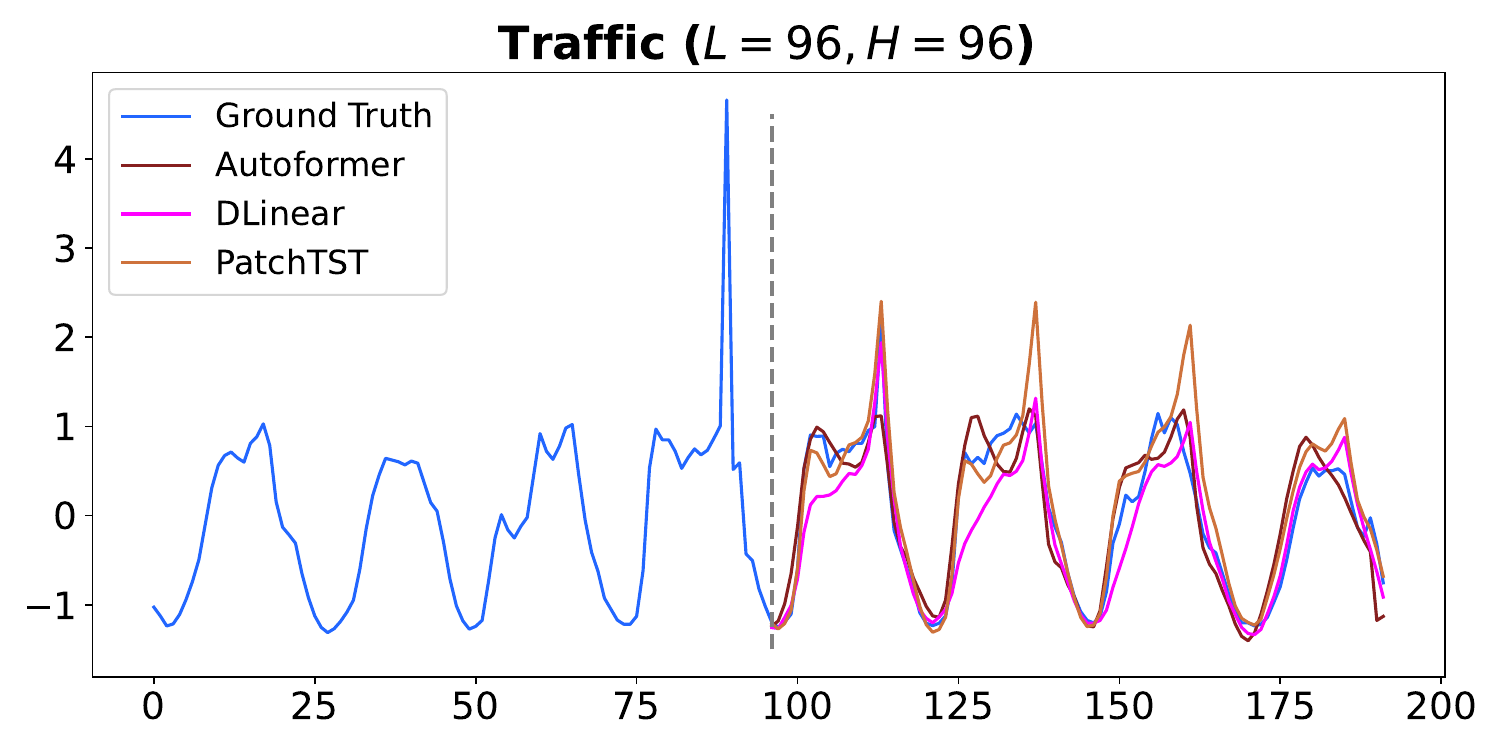}
    }
    \caption{Model predicting visualization of Bi-Mamba+, iTransformer, Autoformer, DLinear and PatchTST on Electricity and Traffic. The look-back window size is set to $L=96$ and the predicting length is set to $H=96$.}
    \label{fig:visual}
\end{figure*}

\subsection{Model Efficiency}
We conduct the following experiments to comprehensively evaluate the model efficiency from (a) predicting accuracy, (b) memory usage and (c) training speed. We set $L=96,H=96$ as the forecasting task and use $Batch=32$ for ETTh1 and Traffic. The results are shown in Fig. \ref{fig:efficiency}(a) and Fig. \ref{fig:efficiency}(b). We freeze the patch length to $P=\frac{1}{4}L$ to provide consistent modeling process. We do not adopt \textbf{Effcient Training}\citep{liu2023itransformer} mechanism for Bi-Mamba+, S-Mamba and iTransformer on Traffic, in order to provide an intuitive comparison of the self-attention mechanism with Mamba.

For small datasets like ETTh1, Bi-Mamba+, S-Mamba, iTransformer, PatchTST and DLinear cost similar memory. For large datasets like Traffic, when the number of variables grows numerous, due to the the quadratic complexity of self-attention, the memory cost by iTransformer increases significantly, while Bi-Mamba+ and S-Mamba retain linear growth of resource usage, benefiting from the linear complexity of SSM. Therefore, Bi-Mamba+ strikes a good balance among predicting performance, training speed and memory usage. 
Overall, the computational cost of Mamba+ is less than self-attention mechanism in Transformer with the same embedding size $D$, which is reflected in the training speed and GPU memory usage. With the better predicting performance, faster training speed and less memory costs, Mamba-like models show great potential in LTSF.

\subsection{Visualization}
We visualize the long-term forecasting results of Bi-Mamba+, PatchTST, DLinear, and Autoformer in Fig. \ref{fig:visual}. We predict 96 steps on Traffic and Electricity. Bi-Mamba+ provides stable prediction results which is the closest to the ground truth.

\section{Conclusion}
In this paper, we propose Bi-Mamba+, a novel model that adaptively captures intra- or inter-series dependencies of MTS data. We propose Mamba+ specifically designed for LTSF. Mamba+ adds a forget gate in Mamba to selectively combine the added new features with the forgotten historical features in a complementary manner, therefore preserving historical information in a longer range. We design a bidirectional structure to model the MTS data more comprehensively. We divide the time series into patches to encourage the model to be focused on the inner evolutionary process or inter-series dependencies at a finer granularity. We explore the validity of Mamba-like structures and find that Mamba has great potential in time series forecasting tasks. Extensive experiments show the superior performance and high efficiency of our model compared to the SOTA methods. Future research will explore more diverse and complex scenarios such as network flow forecasting.

\newpage


\newpage
\appendix
\section{Appendix}
\subsection{Full Results of Ablation Study}
Table \ref{tab:result ablation total} shows the detailed MSE and MAE results of (a) w/o SRA-I which use channel-independent strategy only, (b) w/o SRA-M which use channel-mixing strategy only, (c) w/o Bi which use forward direction Mamba block only, (d) w/o Residual that removes the residual connection, (e) S-Mamba and (f) PatchTST. The look-back window size is set to $L=96$, we conduct experiments on varying predicting length $H\in\{96,192,336,720\}$.

\begin{table*}[t]
\centering
\caption{Full results of ablation studies of (a) w/o SRA-I which use channel-independent strategy only, (b) w/o SRA-M which use channel-mixing strategy only, (c) w/o Bi which use forward direction Mamba block only, (d) w/o Residual that removes the residual connection, (e) S-Mamba and (f) PatchTST.}
\label{tab:result ablation total}
\renewcommand{\arraystretch}{1.2}
\resizebox{0.8\linewidth}{!}{
    \begin{tabular}{cc|c|cc|cc|cc|cc|cc|cc|ccc}
        \cline{2-17}
        &\multicolumn{2}{c|}{Models}& \multicolumn{2}{c|}{w/o SRA-I}& \multicolumn{2}{c|}{w/o SRA-M}& \multicolumn{2}{c|}{w/o Bi}& \multicolumn{2}{c|}{w/o Residual}&\multicolumn{2}{c|}{w/o M \& w/ A} & \multicolumn{2}{c|}{S-Mamba}& \multicolumn{2}{c}{PatchTST}\\
        \cline{2-17}
        &\multicolumn{2}{c|}{Metric}&MSE&MAE&MSE&MAE&MSE&MAE&MSE&MAE&MSE&MAE&MSE&MAE&MSE&MAE\\
        \cline{2-17}
        &\multirow{4}*{\rotatebox{90}{Weather}}& 96    
        & 0.164 & 0.211 & 0.159 & 0.205 & 0.169 & 0.215 & 0.166 & 0.216 & 0.174 & 0.215 & 0.166 & 0.211 & 0.178 & 0.219 \\
        &\multicolumn{1}{c|}{}& 192   
        & 0.219 & 0.255 & 0.205 & 0.249 & 0.212 & 0.257 & 0.214 & 0.260 & 0.222 & 0.260 & 0.216 & 0.253 & 0.224 & 0.259 \\
        &\multicolumn{1}{c|}{}& 336   
        & 0.274 & 0.295 & 0.264 & 0.291 & 0.273 & 0.300 & 0.274 & 0.301 & 0.279 & 0.300 & 0.275 & 0.298 & 0.278 & 0.298 \\
        &\multicolumn{1}{c|}{}& 720   
        & 0.352 & 0.347 & 0.343 & 0.344 & 0.349 & 0.350 & 0.352 & 0.353 & 0.352 & 0.347 & 0.353 & 0.349 & 0.354 & 0.348 \\
        \cline{2-17}
        &\multirow{4}*{\rotatebox{90}{Traffic}}& 96    
        & 0.429 & 0.276 & 0.375 & 0.258 & 0.406 & 0.279 & 0.409 & 0.291 & 0.390 & 0.266 & 0.381 & 0.261 & 0.457 & 0.295 \\
        &\multicolumn{1}{c|}{} & 192   
        & 0.440 & 0.280 & 0.394 & 0.269 & 0.422 & 0.283 & 0.420 & 0.295 & 0.411 & 0.280 & 0.397 & 0.267 & 0.471 & 0.299 \\
        &\multicolumn{1}{c|}{}& 336   
        & 0.453 & 0.286 & 0.406 & 0.274 & 0.439 & 0.289 & 0.449 & 0.315 & 0.421 & 0.282 & 0.423 & 0.276 & 0.482 & 0.304 \\
        &\multicolumn{1}{c|}{}& 720   
        & 0.488 & 0.305 & 0.440 & 0.288 & 0.472 & 0.306 & 0.478 & 0.321 & 0.447 & 0.297 & 0.460 & 0.298 & 0.514 & 0.322 \\
        \cline{2-17}
        &\multirow{4}*{\rotatebox{90}{Electricity}}& 96    
        & 0.163 & 0.250 & 0.140 & 0.238 & 0.149 & 0.245 & 0.148 & 0.250 & 0.148 & 0.243 & 0.142 & 0.238 & 0.174 & 0.259 \\
        &\multicolumn{1}{c|}{}& 192   
        & 0.171 & 0.259 & 0.155 & 0.253 & 0.163 & 0.258 & 0.167 & 0.266 & 0.162 & 0.256 & 0.163 & 0.261 & 0.178 & 0.265 \\
        &\multicolumn{1}{c|}{}& 336   
        & 0.188 & 0.276 & 0.170 & 0.269 & 0.182 & 0.278 & 0.183 & 0.284 & 0.177 & 0.273 & 0.178 & 0.275 & 0.196 & 0.282 \\
        &\multicolumn{1}{c|}{}& 720   
        & 0.229 & 0.311 & 0.197 & 0.293 & 0.208 & 0.302 & 0.219 & 0.316 & 0.210 & 0.303 & 0.207 & 0.303 & 0.237 & 0.316 \\
        \cline{2-17}
        &\multirow{4}*{\rotatebox{90}{ETTh1}}& 96    
        & 0.378 & 0.395 & 0.387 & 0.407 & 0.383 & 0.403 & 0.386 & 0.406 & 0.394 & 0.408 & 0.386 & 0.406 & 0.393 & 0.408 \\
        &\multicolumn{1}{c|}{}& 192   
        & 0.427 & 0.428 & 0.438 & 0.428 & 0.431 & 0.430 & 0.442 & 0.434 & 0.448 & 0.439 & 0.448 & 0.444 & 0.445 & 0.434 \\
        &\multicolumn{1}{c|}{}& 336   
        & 0.471 & 0.445 & 0.486 & 0.446 & 0.474 & 0.450 & 0.483 & 0.457 & 0.458 & 0.449 & 0.494 & 0.468 & 0.474 & 0.451 \\
        &\multicolumn{1}{c|}{}& 720   
        & 0.470 & 0.457 & 0.478 & 0.466 & 0.482 & 0.465 & 0.511 & 0.488 & 0.485 & 0.484 & 0.493 & 0.488 & 0.480 & 0.471 \\
        \cline{2-17}
        &\multirow{4}*{\rotatebox{90}{ETTh2}}& 96    
        & 0.291 & 0.342 & 0.291 & 0.344 & 0.299 & 0.353 & 0.298 & 0.349 & 0.296 & 0.346 & 0.298 & 0.349 & 0.302 & 0.348 \\
        &\multicolumn{1}{c|}{}& 192   
        & 0.368 & 0.392 & 0.371 & 0.393 & 0.374 & 0.400 & 0.382 & 0.400 & 0.377 & 0.393 & 0.379 & 0.398 & 0.388 & 0.400 \\
        &\multicolumn{1}{c|}{}& 336   
        & 0.407 & 0.424 & 0.409 & 0.429 & 0.410 & 0.428 & 0.420 & 0.431 & 0.401 & 0.424 & 0.417 & 0.432 & 0.426 & 0.433 \\
        &\multicolumn{1}{c|}{}& 720   
        & 0.421 & 0.439 & 0.415 & 0.442 & 0.421 & 0.442 & 0.427 & 0.446 & 0.421 & 0.444 & 0.431 & 0.449 & 0.431 & 0.446 \\
        \cline{2-17}
        &\multirow{4}*{\rotatebox{90}{ETTm1}}& 96    
        & 0.320 & 0.360 & 0.320 & 0.363 & 0.323 & 0.363 & 0.326 & 0.365 & 0.317 & 0.357 & 0.331 & 0.368 & 0.329 & 0.367 \\
        &\multicolumn{1}{c|}{}& 192   
        & 0.361 & 0.383 & 0.370 & 0.389 & 0.369 & 0.389 & 0.366 & 0.388 & 0.363 & 0.383 & 0.371 & 0.387 & 0.367 & 0.385 \\
        &\multicolumn{1}{c|}{}& 336   
        & 0.386 & 0.402 & 0.402 & 0.413 & 0.396 & 0.408 & 0.400 & 0.409 & 0.392 & 0.404 & 0.417 & 0.418 & 0.399 & 0.410 \\
        &\multicolumn{1}{c|}{}& 720   
        & 0.445 & 0.437 & 0.463 & 0.451 & 0.457 & 0.443 & 0.452 & 0.443 & 0.451 & 0.438 & 0.471 & 0.448 & 0.454 & 0.439 \\
        \cline{2-17}
        &\multirow{4}*{\rotatebox{90}{ETTm2}} & 96    
        & 0.176 & 0.263 & 0.176 & 0.265 & 0.177 & 0.264 & 0.181 & 0.271 & 0.180 & 0.265 & 0.179 & 0.263 & 0.175 & 0.259 \\
        &\multicolumn{1}{c|}{}& 192   
        & 0.242 & 0.304 & 0.244 & 0.308 & 0.247 & 0.310 & 0.249 & 0.315 & 0.249 & 0.310 & 0.253 & 0.310 & 0.241 & 0.302 \\
        &\multicolumn{1}{c|}{}& 336   
        & 0.304 & 0.344 & 0.309 & 0.348 & 0.314 & 0.355 & 0.314 & 0.354 & 0.310 & 0.349 & 0.312 & 0.349 & 0.305 & 0.343 \\
        &\multicolumn{1}{c|}{}& 720   
        & 0.402 & 0.402 & 0.405 & 0.406 & 0.412 & 0.411 & 0.418 & 0.415 & 0.410 & 0.409 & 0.412 & 0.408 & 0.402 & 0.400 \\
        \cline{2-17}
        &\multirow{4}*{\rotatebox{90}{Solar}}& 96    
        & 0.204 & 0.243 & 0.184 & 0.222 & 0.197 & 0.226 & 0.195 & 0.231 & 0.193 & 0.242 & 0.205 & 0.241 & 0.234 & 0.286 \\
        &\multicolumn{1}{c|}{}& 192   
        & 0.236 & 0.264 & 0.225 & 0.254 & 0.230 & 0.254 & 0.227 & 0.259 & 0.247 & 0.279 & 0.237 & 0.270 & 0.267 & 0.310 \\
        &\multicolumn{1}{c|}{}& 336   
        & 0.252 & 0.270 & 0.249 & 0.270 & 0.260 & 0.288 & 0.246 & 0.276 & 0.268 & 0.299 & 0.256 & 0.287 & 0.290 & 0.315 \\
        &\multicolumn{1}{c|}{}& 720   
        & 0.251 & 0.274 & 0.250 & 0.273 & 0.251 & 0.278 & 0.261 & 0.286 & 0.257 & 0.283 & 0.259 & 0.285 & 0.289 & 0.317 \\
        \cline{2-17}
    \end{tabular}
}
\end{table*}

\subsection{Reprogrammed Back Propagation}
For the original Mamba block, the forward propagation process passes the input sequence through two branches, denoted as $b_1$ and $b_2$.
Mamba calculates the result by the formula $\mathrm{output}=\mathrm{Linear^{y^\prime}(y\otimes SiLU(z))}$, where $\mathrm{y}$ is the output of SSM in $b_1$ and $\mathrm{z}$ is the linear layer output in $b_2$. Our proposed Mamba+ block adjust the calculation process to $\mathrm{output}=\mathrm{Linear^{y^\prime}\left(y\otimes SiLU(z)+x^\prime\otimes(1-\sigma(z))\right)}$, where $x^\prime$ is the output of 1-D convolutional layer in $b_1$. The hardware-aware parallel computing algorithm of Mamba is implemented through CUDA programming. We modify the corresponding \textbf{\textit{.cuh}} files programmed with \textbf{\textit{C++}} and implement the back propagation logic in Alg. \ref{alg:alg4}. We obtain the \textit{selective\_scan\_cuda} program by enforcing the use of source code compilation, so that the training process of the model is completed by Mamba+.

\begin{algorithm}[h]
\caption{The back propagation process adjustment of Mamba+ block}\label{alg:alg4}
\renewcommand{\algorithmicrequire}{\textbf{Input:}}
\renewcommand{\algorithmicensure}{\textbf{Output:}}
\begin{algorithmic}[1]
\REQUIRE Loss gradient $\frac{\delta\mathcal{L}}{\delta\mathrm{output}}$ of linear layer $\mathrm{Linear^{y^\prime}}$
\ENSURE Gradient $\frac{\delta\mathcal{L}}{\delta\mathrm{y}}$ passed to SSM module in $b_1$, gradient $\frac{\delta\mathcal{L}}{\delta\mathrm{x^\prime}}$ passed to 1-D convolutional layer in $b_1$ and gradient $\frac{\delta\mathcal{L}}{\mathrm{z}}$ passed to $\mathrm{z}$ in $b_2$.
\STATE $\frac{\delta\mathrm{output}}{\delta\mathrm{y}}=\mathrm{SiLU(z)}=\mathrm{\frac{z}{1+\exp(-z)}}$
\STATE Store: $\frac{\delta\mathcal{L}}{\delta\mathrm{y}}=\frac{\delta\mathcal{L}}{\delta\mathrm{output}}\cdot\frac{\delta\mathrm{output}}{\delta\mathrm{y}}$
\STATE $\frac{\delta\mathrm{output}}{\delta\mathrm{x^\prime}}=\mathrm{1-\sigma(z)}=\mathrm{\frac{\exp(-z)}{1+\exp(-z)}}$
\STATE Store: $\frac{\delta\mathcal{L}}{\delta\mathrm{x^\prime}}=\frac{\delta\mathcal{L}}{\delta\mathrm{output}}\cdot\frac{\delta\mathrm{output}}{\delta\mathrm{x^\prime}}$
\STATE $\frac{\delta\mathrm{output}}{\delta\mathrm{z}}=\mathrm{y\cdot\frac{\delta SiLU(z)}{\delta z}+x^\prime\cdot\frac{\delta(1-\sigma(z))}{\delta z}}$
\STATE $\frac{\delta\mathrm{output}}{\delta\mathrm{z}}=\mathrm{y\cdot\frac{1}{1+\exp(-z)}\left(1+z(\frac{\exp(-z)}{1+\exp(-z)})\right)-}$ \\ 
\quad\quad\quad\quad\;$\mathrm{x^\prime\cdot\frac{1}{1+\exp(-z)}\left(\frac{\exp(-z)}{1+\exp(-z)}\right)}$
\STATE Store: $\frac{\delta\mathcal{L}}{\delta\mathrm{z}}=\frac{\delta\mathcal{L}}{\delta\mathrm{output}}\cdot\frac{\delta\mathrm{output}}{\delta\mathrm{z}}$
\STATE \textbf{return} $\frac{\delta\mathcal{L}}{\delta\mathrm{y}}$, $\frac{\delta\mathcal{L}}{\delta\mathrm{x^\prime}}$, $\frac{\delta\mathcal{L}}{\delta\mathrm{z}}$
\end{algorithmic}
\label{alg1}
\end{algorithm}

\end{document}